\documentclass{article}

\usepackage{PRIMEarxiv}

\usepackage[utf8]{inputenc} 
\usepackage[T1]{fontenc}    
\usepackage{hyperref}       
\usepackage{url}            
\usepackage{booktabs}       
\usepackage{amsfonts}       
\usepackage{nicefrac}       
\usepackage{subcaption}
\usepackage{microtype}      
\usepackage{lipsum}
\usepackage{amsmath}
\usepackage{bm}
\usepackage{fancyhdr}       
\usepackage{graphicx}       
\usepackage{hyperref}
\usepackage{cleveref}
\usepackage{amsthm}
\newtheorem{finding}{Finding}
\graphicspath{{media/}}     
\usepackage{wrapfig}
\title{Vision Transformers that Never Stop Learning
\thanks{\textit{\underline{Citation}}: 
\textbf{Authors. Title. Pages.... DOI:000000/11111.}} 
}

\author{
Caihao Sun$^{1}$,
Minqi Yuan$^{2\dagger}$
Shiyuan Wang$^{3}$
Jiayu Chen$^{14\dagger}$\\[4pt]
$^1$University of Hong Kong \quad
$^2$The Hong Kong Polytechnic University\\
$^3$Technische Universität Dresden \quad
$^4$INFIFORCE\\[6pt]
\small
$\dagger$Co-corresponding author
}

\begin{document}
\maketitle
\begin{abstract}
Loss of plasticity refers to the progressive inability of a model to adapt to new tasks and poses a fundamental challenge for continual learning. While this phenomenon has been extensively studied in homogeneous neural architectures, such as multilayer perceptrons, its mechanisms in structurally heterogeneous, attention-based models such as Vision Transformers (ViTs) remain underexplored. In this work, we present a systematic investigation of loss of plasticity in ViTs, including a fine-grained diagnosis using local metrics that capture parameter diversity and utilization. Our analysis reveals that stacked attention modules exhibit increasing instability that exacerbates plasticity loss, while feed-forward network modules suffer even more pronounced degradation. Furthermore, we evaluate several approaches for mitigating plasticity loss. The results indicate that methods based on parameter re-initialization fail to recover plasticity in ViTs, whereas approaches that explicitly regulate the update process are more effective. Motivated by this insight, we propose ARROW, a geometry-aware optimizer that preserves plasticity by adaptively reshaping gradient directions using an online curvature estimate for the attention module. Extensive experiments show that ARROW effectively improves plasticity and maintains better performance on newly encountered tasks.
\end{abstract}

\section{Introduction}\label{sec:intro}
The development of learning systems that can continuously adapt to new tasks while retaining previously acquired knowledge is a cornerstone of achieving artificial general intelligence (AGI) \cite{wang2024comprehensive,dohare2024loss,yuan2025plasticine}. This paradigm, known as continual learning, has been attracting significant attention across the AI community. Unlike conventional deep learning paradigms that typically rely on fixed, independent, and identically distributed datasets, continual learning seeks to optimize models over a non-stationary stream of tasks where data distributions shift over time \cite{li2026continualpolicydistillationdistributed,dohare2024loss}. However, a significant challenge to this objective is the ``\textbf{loss of plasticity}'', a phenomenon in which models progressively lose their ability to learn new concepts as training continues \cite{sokar2023dormant}. 

Recent studies have increasingly focused on structural and optimization-related factors underlying plasticity loss across various learning paradigms. In supervised class-incremental learning, \cite{kim2023stability} demonstrates that many existing algorithms are heavily biased toward stability, which significantly reduces the model's capacity for future learning. This degradation is further formalized by \cite{dohare2024loss}, who show that standard deep learning methods inevitably lose plasticity in long-term continual settings. To address this, \cite{dohare2024loss} proposes a continual backpropagation (CBP) approach to inject plasticity through unit replacement. In reinforcement learning (RL), \cite{sokar2023dormant} identifies that neurons become progressively inactive and redundant, and introduces the ReDo algorithm to recycle these units and maintain network expressivity. Furthermore, \cite{muppidi2024fast} studies the plasticity problem from an optimization perspective and proposes TRAC, a parameter-free optimizer that mitigates plasticity loss in non-stationary RL environments by adaptively adjusting learning steps based on online convex optimization principles. 

While these existing studies provide foundational insights, they primarily focus on homogeneous or simpler architectures such as multilayer perceptrons (MLPs) or convolutional neural networks (CNNs). In contrast, the Transformer architecture \cite{vaswani2017attention} has become the backbone of modern large-scale foundation models. However, their behavior regarding the plasticity loss remains significantly underexplored, particularly for Vision Transformers (ViTs) \cite{dosovitskiy2020image}.Given that ViTs underpin many current computer vision systems, understanding their lifelong learning capabilities is both critical and urgently needed \cite{wang2022continual}.

\begin{figure}[t!]
    \centering
    \begin{minipage}[t]{0.55\textwidth}
        \centering
        \includegraphics[width=\linewidth]{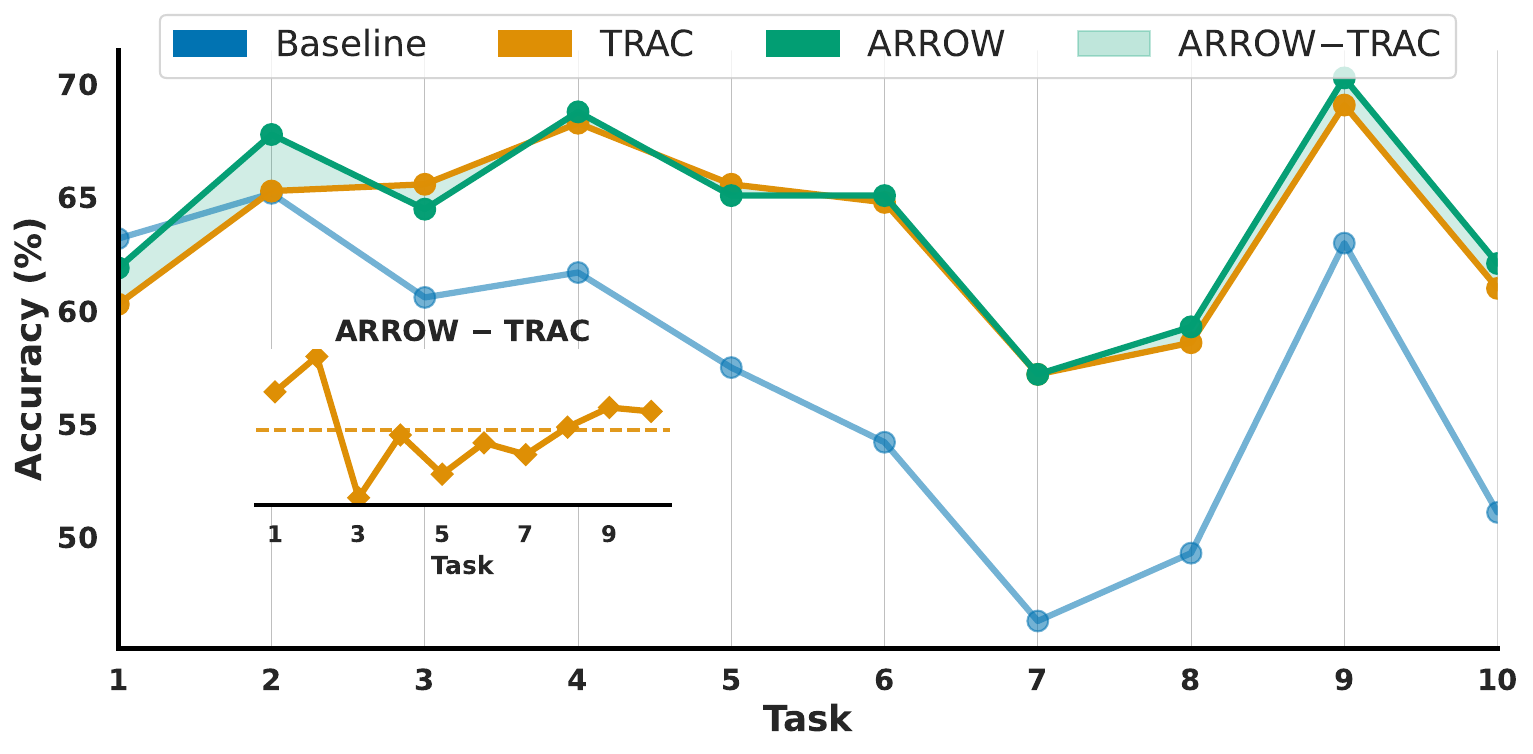}
        \subcaption{}
        \label{fig:prefix_aat}
    \end{minipage}
    \hfill
    \begin{minipage}[t]{0.35\textwidth}
        \centering
        \includegraphics[width=\linewidth]{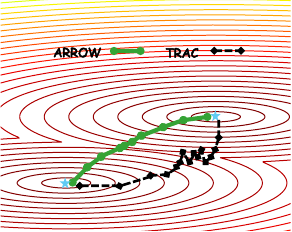}
        \subcaption{}
        \label{fig:prefix_arrwo_workflow}
    \end{minipage}
    \caption{(a) ViTs exhibit plasticity loss in task-incremental CIFAR-100, with AAT degrading over the task stream. (b) ARROW, a geometry-aware optimizer with an online low-rank curvature estimate, stabilizes optimization rather than exhibits oscillatory updates under distribution shift. Both ARROW and TRAC outperform the vanilla ViT, and ARROW’s advantage over TRAC increases in later tasks (inset).}
\end{figure}
Inspired by the discussions above, we conduct a systematic study of ViTs' plasticity in continual learning settings. Our main contributions are threefold:
\begin{itemize}
    \item Through comprehensive diagnostic experiments on long-term non-stationary task streams, we identify that ViTs suffer from a unique form of plasticity loss. This is indicated by a rapid collapse in the effective rank and a significant increase in dormant units within the feed-forward network modules. 
    \item We evaluate a series of existing strategies for mitigating plasticity loss, including structural re-initialization and optimization-based methods. Experiments demonstrate that they are insufficient for maintaining the representational diversity required by the complex, multimodal landscape of ViTs.
    \item To address these structural and geometric issues, we propose \textbf{ARROW}: \textbf{A}daptive \textbf{R}ank-\textbf{R}eshaping via \textbf{O}nline \textbf{W}indowed covariance, a novel geometry-aware optimizer that utilizes a low-rank curvature proxy to adaptively reshape gradient updates, thereby amplifying neglected directions and maintaining representational dimensionality throughout the learning process.
\end{itemize}

\section{Related Work}\label{sec:related work}
\subsection{Continual Learning}
Continual learning seeks to enable models to learn from non-stationary data distributions while mitigating the interference between successive tasks \cite{wang2024comprehensive}. For instance, continual supervised learning aims to retain knowledge across tasks in static datasets \cite{cha2021co2l,davari2022probing}, while continual RL \cite{sokar2023dormant,muppidi2024fast} aims to build agents that adapt to dynamic environments while maintaining prior experience. The primary challenge in CL is governed by the stability–plasticity dilemma \cite{kim2023stability}. Stability characterizes the ability to preserve past knowledge and prevent catastrophic forgetting, which has been extensively studied through regularization-based \cite{kirkpatrick2017overcoming,ritter2018online,schwarz2018progress}, replay-based \cite{caccia2020online,wang2022memory,kumari2022retrospective}, and optimization-based approaches \cite{wang2021training,lin2022trgp,kao2021natural}, etc. In contrast, plasticity refers to the model's capacity to adapt to new information \cite{wang2024comprehensive}, which determines the upper bound of a model's lifelong performance. In this paper, we investigate the plasticity within the structural context of ViTs, an area that remains significantly underexplored.

\subsection{Plasticity Loss Mitigation in Continual Learning}
The phenomenon of plasticity loss refers to the model gradually losing its ability to adapt or effectively update its parameters \cite{sokar2023dormant}. To mitigate the plasticity loss, existing studies can be broadly categorized into reset-based intervention, normalization, regularization, activation function, and optimizer \cite{yuan2025plasticine}. For instance, Shrink and Perturb (SnP) \cite{ash2020warm} periodically scales the model magnitude and injects noises to restore adaptability. Normalization and Projection (NaP) \cite{lyle2024normalization} introduces additional normalization layers before nonlinear transformations and rescales weights. L2 regularization \cite{lyle2023understanding} constrains L2 norm of parameters through penalty terms. In addition, widely-used ReLU activation function can be replaced with CReLU \cite{abbas2023loss}. Novel optimizers like TRAC \cite{muppidi2024fast} and KRON \cite{castanyer2025stable} are proposed to dynamically regulate updates or incorporate second-order approximation optimization. In this paper, we propose ARROW, a geometry-aware optimizer that reshapes gradient directions using a low-rank curvature approximation.

\subsection{Plasticity in Vision Transformers}
The structural heterogeneity of ViTs introduces challenges for maintaining plasticity compared to homogeneous architectures e.g., MLPs, CNNs. Recent studies have identified distinct behaviors across ViT components: (1) attention modules are prone to early head specialization and gradient inconsistency \cite{liang2025attention, zhou2021deepvit}; (2) feed-forward networks (FFNs) suffer from feature saturation and representation collapse \cite{chen2022principle}. To address these issues, existing research has focused on architectural modifications, such as inter-task contrastive mechanisms \cite{wang2022continual, wang2022online}, attention simplification \cite{bekal2025continual}, and prompt associated with task-specific information \cite{smith2023coda, jiang2025dupt}. Despite these advances, a systematic, layer-wise diagnosis of ViT plasticity remains absent. Current literature predominantly emphasizes the attention architecture, leaving the hierarchical evolution of plasticity largely unexplored. This paper addresses this gap by providing a holistic analysis of how individual components and depth-wise structures contribute to the erosion of learning capacity.

\section{Background}\label{sec:background}
\subsection{Task-Incremental Continual Learning}\label{sec:ticl}
In task-incremental settings, a learner is presented with a stream of $T$ tasks $\{\mathcal{T}_t\}_{t=1}^T$. Each task $\mathcal{T}_t$ is associated with a data distribution $\mathcal{P}_t$ over the input space $\mathcal{X}$ and label space $\mathcal{Y}_t$, from which a finite dataset $\mathcal{D}_t = \{(x_i^t, y_i^t)\}_{i=1}^{n_t}$ is sampled. The objective is to minimize the empirical risk across all tasks encountered:
\begin{equation}
    \min_{\bm\theta} \sum_{t=1}^T \mathbb{E}_{(x,y) \sim \mathcal{D}_t} [ \mathcal{L}(f(x, t; \bm{\theta}), y) ],
\end{equation}
where $f(\cdot, t; \bm{\theta})$ denotes the model parameterized by $\bm{\theta}$. Task-incremental continual learning is commonly adopted for analyzing plasticity, as it isolates plasticity degradation from task-inference errors and provides a pure evaluation of representational adaptability under distribution shift \cite{dohare2024loss}.

\subsection{Architecture of ViTs}
\begin{figure}[t]
\centering
\includegraphics[width=0.7\linewidth]{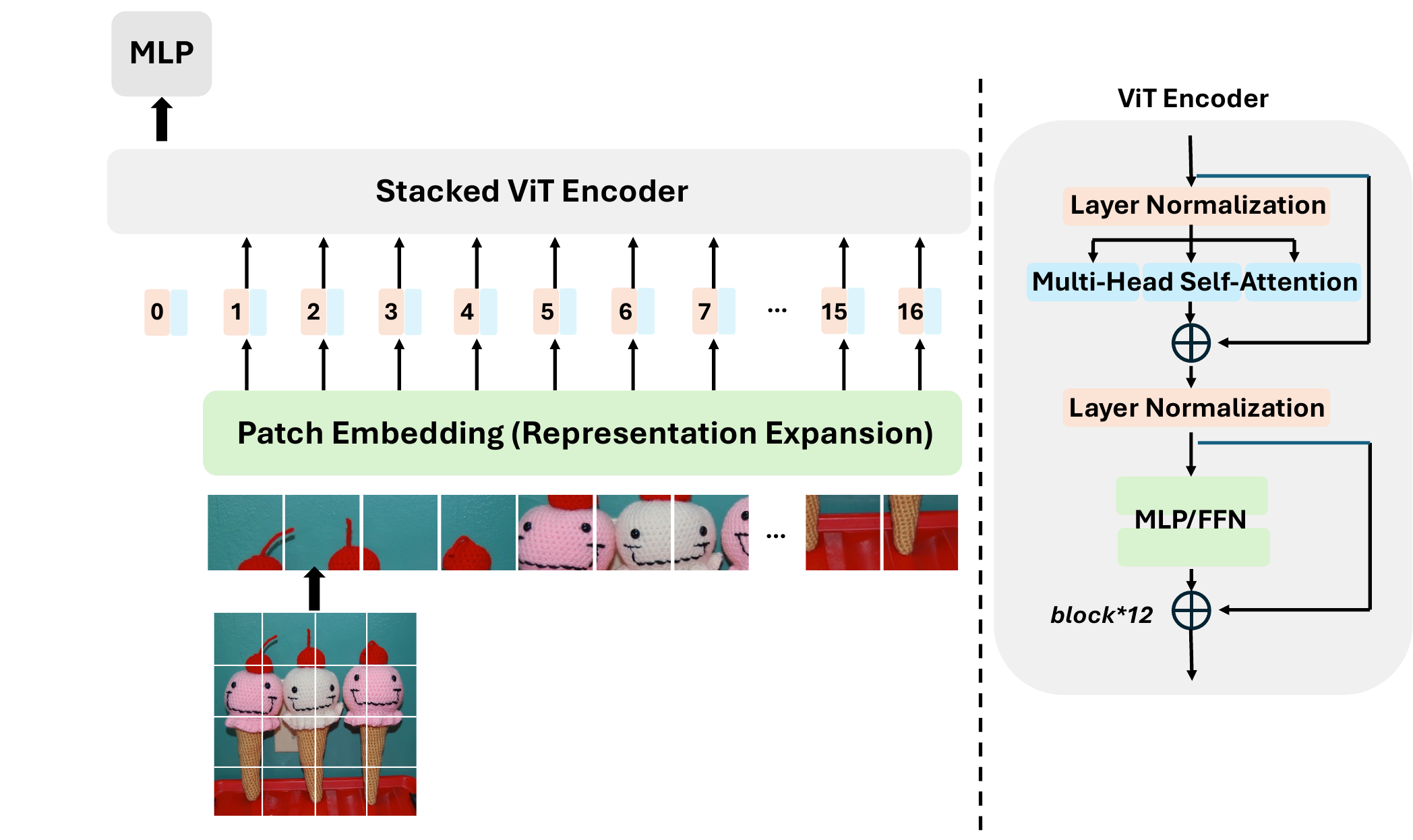}
\caption{Architecture of the ViT.}
\label{vit}
\end{figure}

Given input images $x \in \mathbb{R}^{H \times W \times C}$, Vision Transformers (ViTs) divide images into pathes and projects them into a fixed-dimensional embedding space. A learnable classification token (CLS token) and positional embeddings are added.
As illustrated in Fig.~\ref{vit}, the embedded tokens are then processed by stacked ViT blocks. Each block consists of a Multi-Head Self-Attention (MHSA) module followed by a Feed-Forward Network (FFN), with Layer Normalization and residual connections applied to both sub-modules. The final CLS token representation is forwarded to the classification heads for prediction.

The MHSA module captures global correlations among tokens, while the FFN performs channel-wise nonlinear transformations to enhance representational capacity. Nested learning \cite{behrouz2025nested} suggests that attention modules are associated with high-frequency knowledge adaptation, whereas FFNs may contribute more to long-term representational consolidation.

\subsection{Evaluation Metrics of Plasticity}
To quantitatively characterize the plasticity loss, we adopt a global metric: Average Accuracy across Tasks (AAT) \cite{liang2023adaptive}, and local metrics: effective rank (erank); stable rank (srank); Fraction of Active Units (FAU); and weight magnitude \cite{dohare2024loss}, respectively defined as follows:
\begin{small}
\begin{equation}
\label{eq:metrics_summary}
\renewcommand{\arraystretch}{2.5} 
\setlength{\arraycolsep}{10pt}   
\begin{array}{cc}
\displaystyle \mathrm{AAT} = \frac{1}{T} \sum_{t=1}^{T} \mathrm{Acc}_t & 
\displaystyle \|\mathbf{W}\|_F = \left( \sum_{i,j} W_{ij}^2 \right)^{1/2} \\
\displaystyle \mathrm{srank}(\mathbf{A}) = \sum_{i=1}^{r} \frac{\sigma_i^2}{\sigma_{\max}^2} & 
\displaystyle \mathrm{erank}(\mathbf{A}) = \exp \left( - \sum_{i=1}^r p_i \log p_i \right) \\
\displaystyle \mathrm{FAU}^{(l)} = \frac{1}{N_l} \sum_{i=1}^{N_l} \mathbb{P}_{x \sim \mathcal{D}} ( a_i^{(l)}(x) > 0 ) & 
\displaystyle \mathrm{FDU}^{(l)} = 1 - \mathrm{FAU}^{(l)}
\end{array}
\end{equation}
\end{small}
where $\sigma$ represents the singular values via Singular Value Decomposition (SVD) of matrix $A$, rank $r$. $W$, $F$ are the annotations of weight matrix and Frobenius norm. Higher effective or stable rank reflects a more representation subspace, which is conducive to plasticity adaptation, while excessive growth in weight magnitude suggests parameter rigidity. For the FAU, $a_i^{(l)}(x)$ denotes the activation of the $i^{th}$ unit from layer $l$.

\section{The Loss of Plasticity in ViTs}\label{sec:loss}
\subsection{Phenomenon and Diagnosis}
To systematically investigate the loss of plasticity in ViTs, we conduct a comprehensive diagnostic analysis under a task-incremental continual learning (TIL) paradigm, as introduced in Section~\ref{sec:ticl}. Following the practice in \cite{dohare2024loss} and \cite{sokar2023dormant}, we construct a benchmark using the CIFAR-100 dataset \cite{krizhevsky2009learning}, which forms a long-horizon sequence of 200 tasks (5 classes per task). Throughout this section, we employ the standard vanilla ViT-B/16 \cite{dosovitskiy2020image} as our primary backbone to ensure the generalizability of our findings.

\subsubsection{The Empirical Evidence of Plasticity Loss in ViTs}

To establish a baseline for optimal plasticity, we set an upper bound by reinitializing the model weights and optimizer states at the onset of each new task, thereby effectively bypassing inter-task interference. Furthermore, to isolate the effects of architectural heterogeneity, we introduce two comparative baselines: (i) a standard MLP, and (ii) a one-block ViT. The latter is specifically designed to match the parameter scale of the MLP, enabling a capacity-controlled comparison between the homogeneous MLP structure and the heterogeneous (MHSA + FFN) structure of ViTs while neutralizing the confounding variable of depth.

\begin{figure}[h!]
    \centering
    \begin{minipage}[t]{0.48\textwidth}
        \centering
        \includegraphics[width=\linewidth]{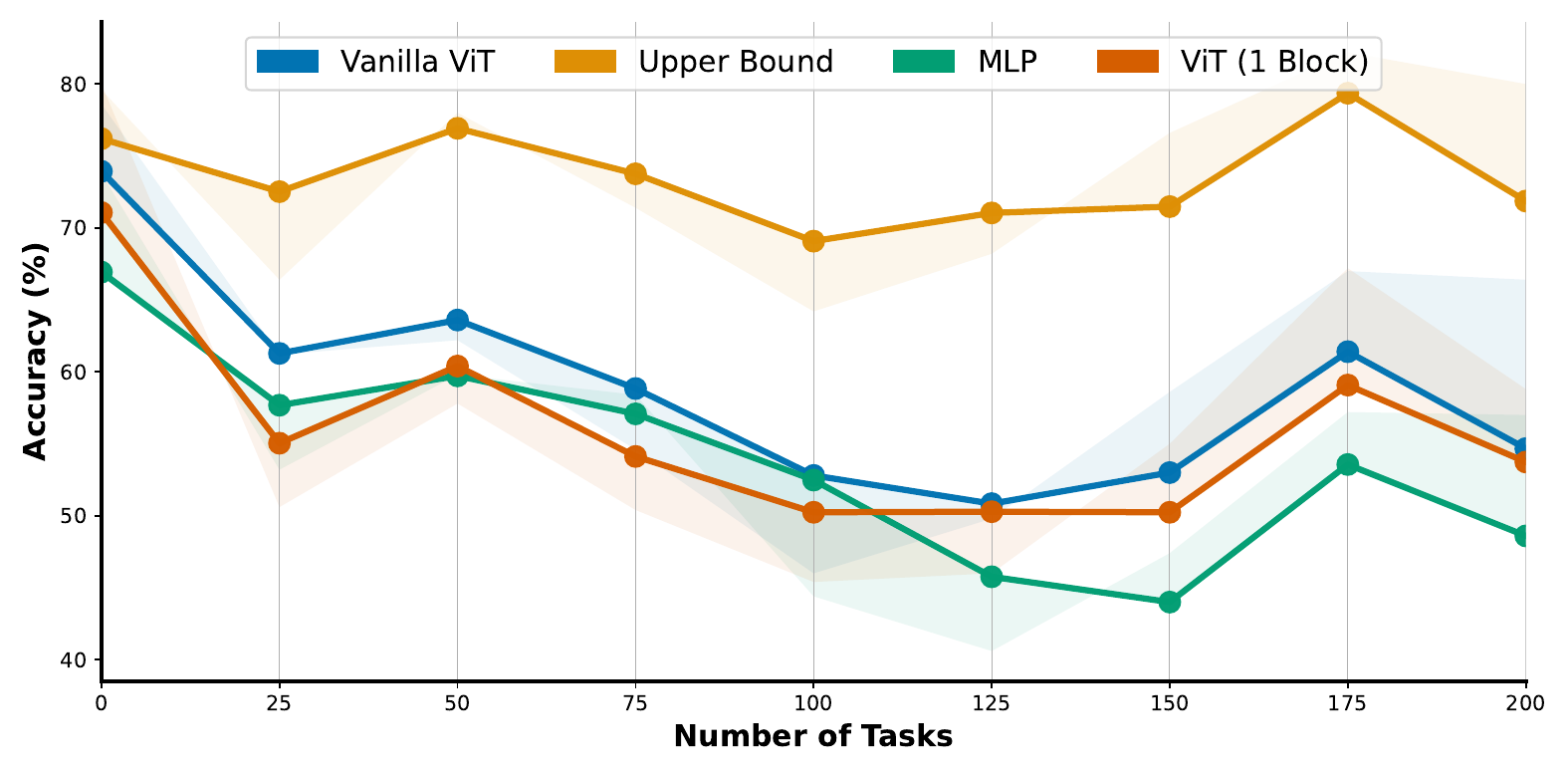}
        \subcaption{}
        \label{fig:aat_comparison_vit_mlp}
    \end{minipage}
    \hfill
    \begin{minipage}[t]{0.48\textwidth}
        \centering
        \includegraphics[width=\linewidth]{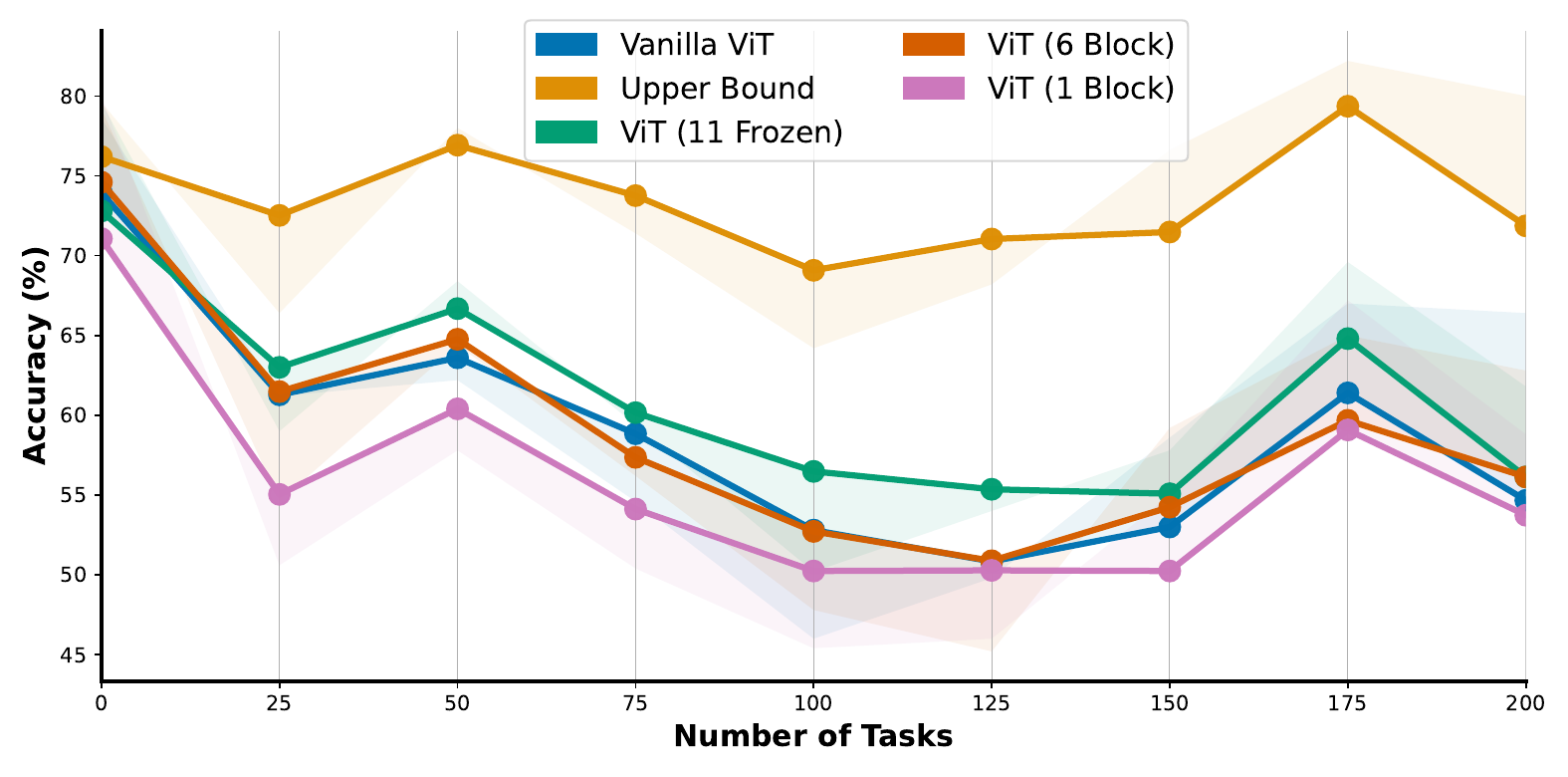}
        \subcaption{}
        \label{fig:aat_comparison_different_vit}
    \end{minipage}
    \caption{Performance comparison regarding the AAT of the MLP and different ViT architectures. All the models exhibit significant performance loss as the task sequence progresses, indicating a broad phenomenon of plasticity loss. The solid line and shaded region represent the mean and standard deviation over multiple runs, respectively.}
\end{figure}

As illustrated in Fig.~\ref{fig:aat_comparison_vit_mlp}, all models exhibit a significant downward trend regarding the AAT metric as the task sequence progresses, demonstrating a broad loss of plasticity. Specifically, the vanilla ViT achieves an AAT of 0.590, resulting in a substantial 14.9\% degradation compared to the upper bound. Notably, both the one-block ViT and the full ViT outperform the MLP model, yet the persistent gap relative to the upper bound confirms that the Transformer architecture also suffers from the plasticity loss. However, the performance advantage of ViT over MLP suggests that its structural heterogeneity may be inherently robust to representation collapse, indicating a more complex underlying mechanism than that observed in standard FFNs.

\begin{finding}
    ViTs exhibit significant plasticity loss during long-term continual learning. Although their heterogeneous architecture offers greater resilience than homogeneous networks (\textit{e.g.}, MLPs), it fails to prevent the progressive deterioration of learning capacity.
\end{finding}

To further investigate how architectural depth and hierarchy affect this phenomenon, we analyze the performance across various ViT configurations: (a) a one-block ViT, (b) a six-block ViT, and (c) a vanilla ViT in which the parameters of the first 11 blocks are frozen during continual training. As shown in Fig.~\ref{fig:aat_comparison_different_vit}, freezing the first 11 blocks of a vanilla ViT effectively improves its performance compared to the fully trainable version, and the one-block ViT performs worst due to limited capacity. These results suggest a hierarchy-dependent loss of plasticity. In ViTs, shallow blocks typically capture low-level features while deeper blocks encode task-specific representations. Freezing early layers helps preserve stable, high-dimensional features and reduces representation drift. In contrast, continuously updating all layers subjects early blocks to conflicting task objectives, forcing gradients into a restricted subspace and accelerating plasticity loss.

\subsubsection{Diagnostic Analysis of MHSA and FFN Modules}
To systematically diagnose the plasticity degradation in ViTs, we conduct a multi-perspective analysis focusing on four key dimensions: (1) \textbf{Representation Dynamics}, tracking the effective rank of feature flows; (2) \textbf{Parameter Geometry}, monitoring weight magnitude and subspace dimensionality; (3) \textbf{Semantic Aggregation}, analyzing the evolution of the CLS token; and (4) \textbf{Neuron Utilization}, quantifying the proportion of dead units in FFNs. Together, these metrics offer a holistic view of how task-sequential learning erodes the model's adaptive capacity.

\begin{figure}[h!]
    \centering
    \begin{minipage}[t]{0.95\linewidth}
        \centering
        \includegraphics[width=\linewidth]{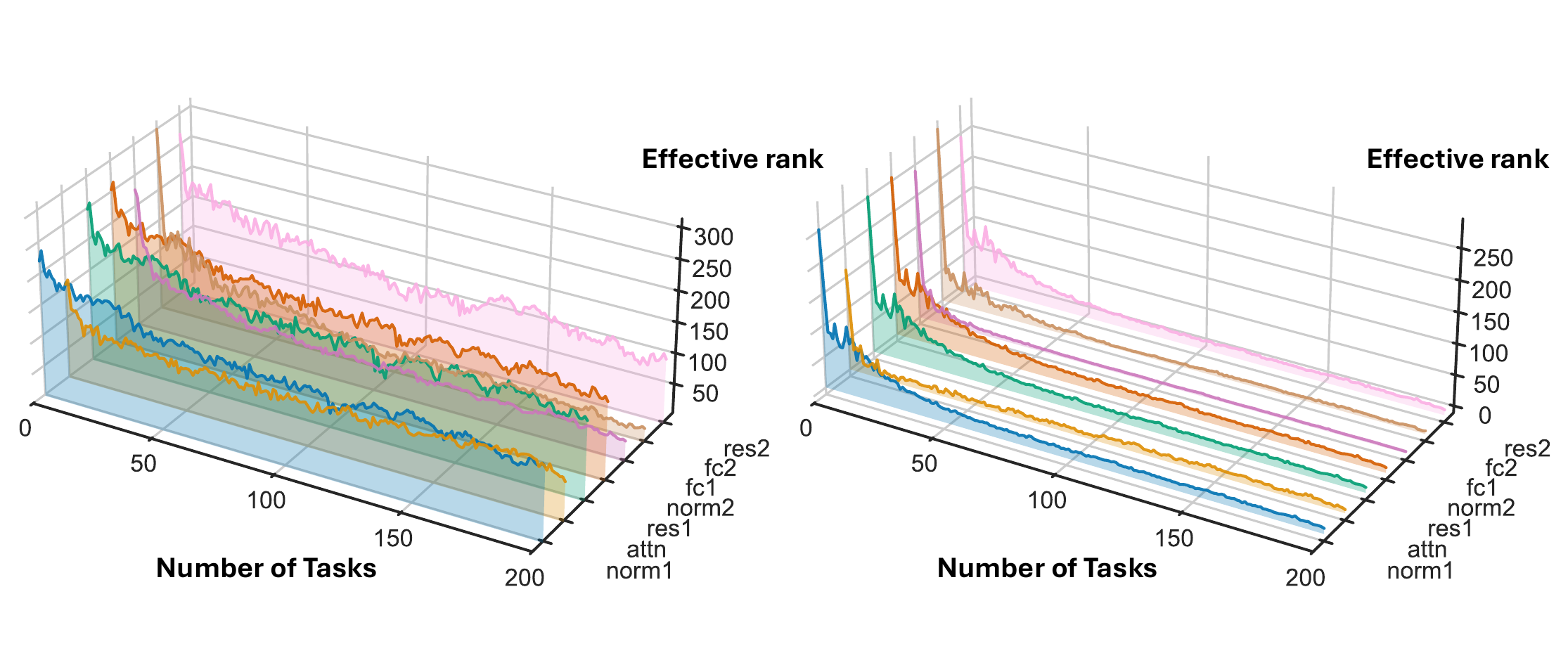}
        \subcaption{}
        \label{fig:erank_blocks_1_8}
    \end{minipage}
    \vfill
    \begin{minipage}[t]{0.95\linewidth}
        \centering
        \includegraphics[width=\linewidth]{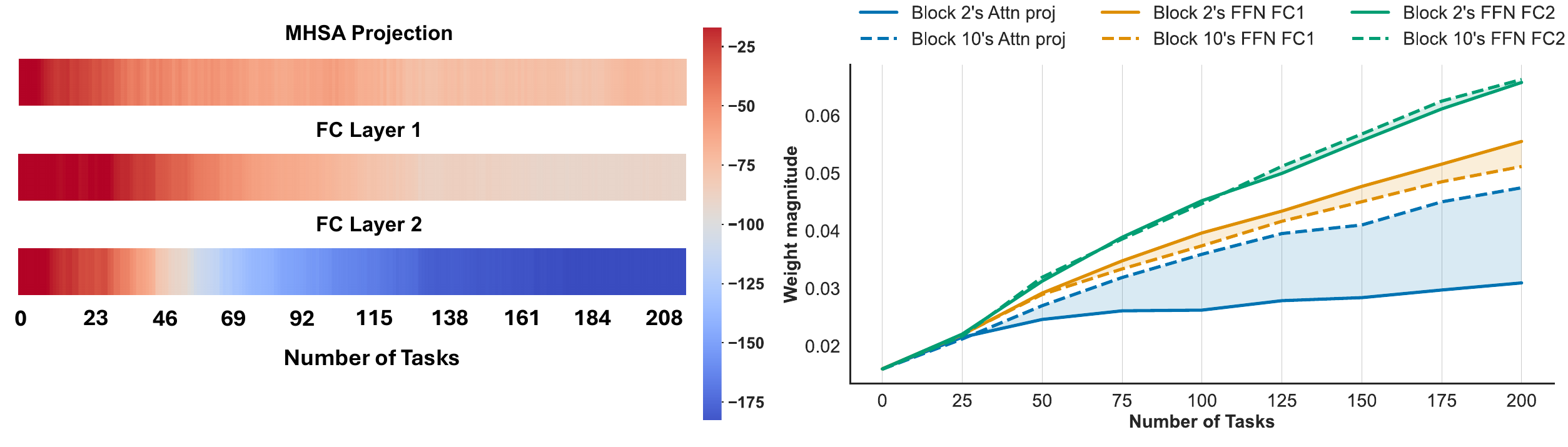}
        \subcaption{}
        \label{fig:weight_magnitude_12}
    \end{minipage}
    \caption{(a) The effective rank of features produced by different components in the 2nd block (left) and the 8th block (right) of the ViT. (b) $\Delta$-heatmap (left) of the effective rank differences between the 1st block and the 12th block in ViT. The right figure shows the weight magnitude variation from the 3rd and 11th blocks.}
\end{figure}

As shown in Fig.~\ref{fig:erank_blocks_1_8}, shallow blocks maintain relatively stable effective rank throughout the task stream, while deeper blocks exhibit rapid and persistent subspace contraction. This collapse is also observed in the weight space, as shown in Fig.~\ref{fig:weight_magnitude_12}, where weight magnitudes in deeper layers grow significantly faster than those in shallow ones, suggesting that parameter consolidation and rigidity intensify as information flows toward the output. This hierarchical stacking enhances representation but exacerbates plasticity loss.

\begin{figure}[h!]
    \centering
    \begin{minipage}[t]{\linewidth}
        \centering
        \includegraphics[width=0.6\linewidth]{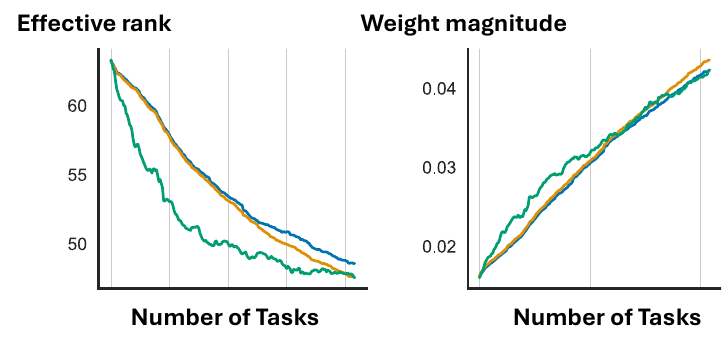}
        \subcaption{}
        \label{fig:erank_qkv}
    \end{minipage}
    \vfill
    \begin{minipage}[t]{\linewidth}
        \centering
        \includegraphics[width=0.9\linewidth]{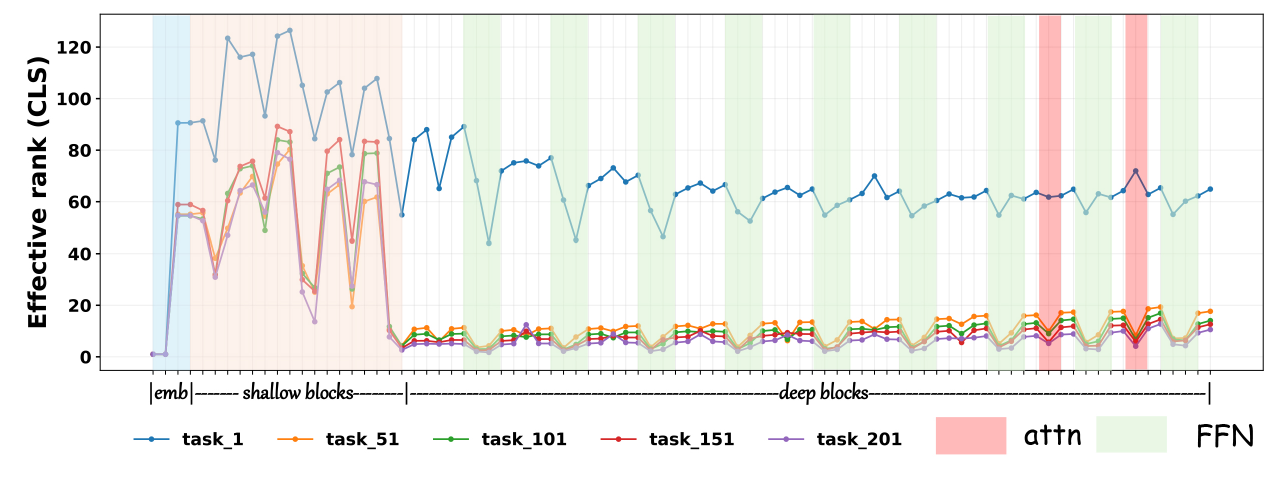}
        \subcaption{}
        \label{fig:erank_cls_token}
    \end{minipage}
    \caption{(a) The effective rank (left) and weight magnitude (right) of one of the attention heads. The green curve represents the $V$ matrix, which is most unstable. The blue and red curves represent the $Q$ and $K$, respectively. (b) The effective rank of CLS token across sampled tasks. The values on embedding layers, shallow and deep blocks are tracked sequentially. Detailed component labels of ViT are provided in the Appendix.}
\end{figure}

Our analysis also reveals distinct degradation patterns across ViT components, highlighting a fundamental divergence in how different modules lose plasticity. Notably, the FFNs emerge as the structural bottleneck, exhibit significantly lower effective rank, and more aggressive weight growth than attention modules across all depths. This vulnerability is further substantiated by the neuron-level metrics in the Appendix, where a progressive accumulation of FDU suggests that FFNs undergo a severe loss of expressivity as training proceeds. In contrast, attention modules demonstrate a more resilient yet depth-sensitive behavior. While stable in shallow layers, attention modules become increasingly unstable in deeper blocks. Notably, as shown in Fig.~\ref{fig:erank_qkv}, the value (V) matrix exhibits greater instability than query (Q) or key (K) matrices, indicating that the content projection is more susceptible to task shifts than the addressing mechanism.

Furthermore, the CLS token tracks these internal dynamics due to its role as the primary semantic aggregator. As shown in Fig.~\ref{fig:erank_cls_token}, while patch embeddings initially expand the representation space, this diversity diminishes as the token propagates through deeper layers. This global decay mirrors our modular findings: the CLS token largely maintains its rank in shallow blocks but suffers significant contraction within FFNs and deeper attention layers. 

\begin{finding}
    The loss of plasticity in ViTs is both depth-amplified and module-dependent. While degradation intensifies in deeper layers, it is driven by the structural collapse of FFNs. In contrast, attention modules maintain relative stability in shallow blocks but become increasingly unstable as the hierarchy deepens, failing to compensate for the overall representational decay.
\end{finding}

\begin{figure}[h!]
    \centering
    \begin{minipage}[t]{0.48\textwidth}
        \centering
        \includegraphics[width=\linewidth]{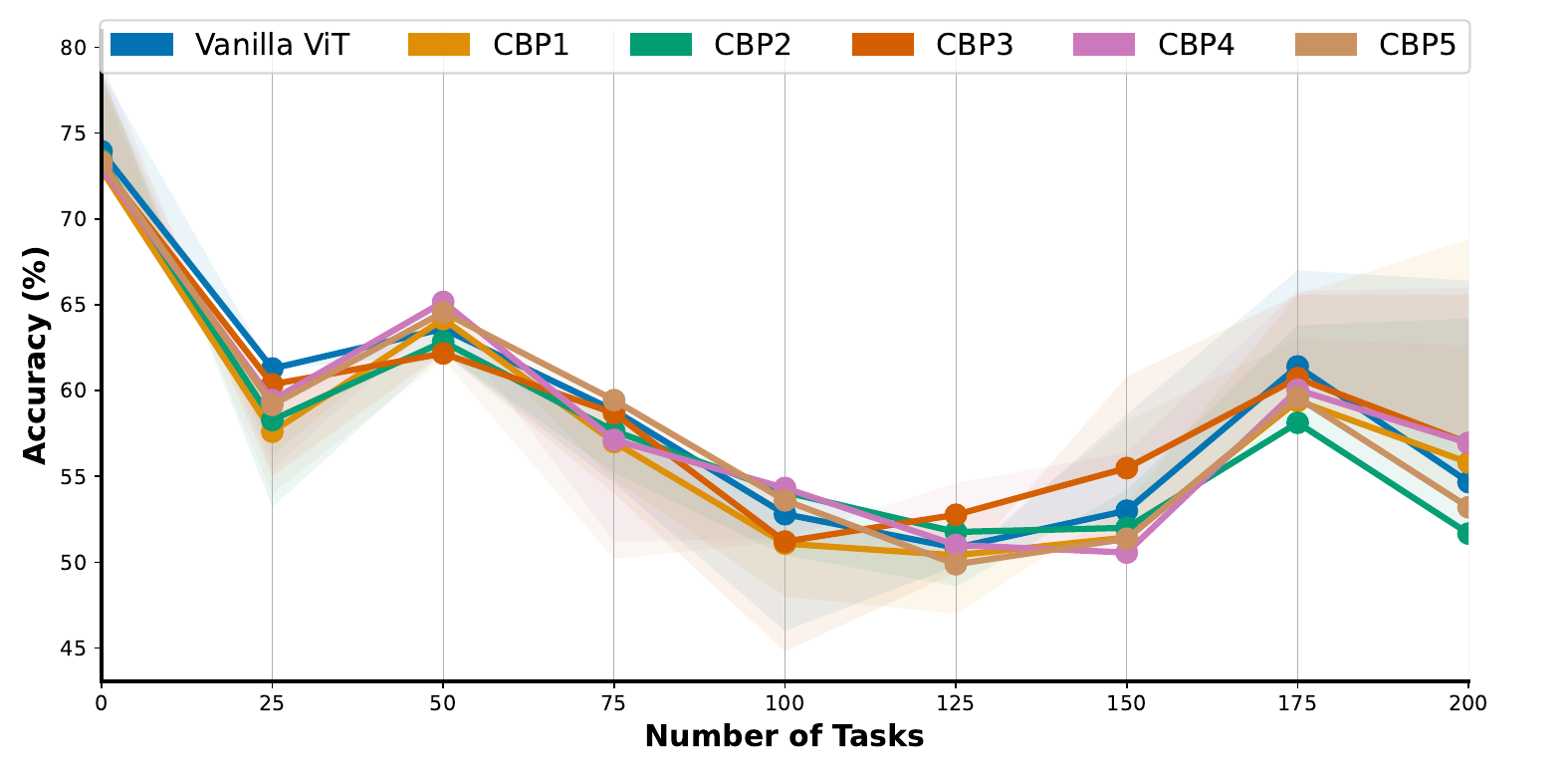}
        \subcaption{}
        \label{fig:cbp_accuracies_comparison_moving_average}
    \end{minipage}
    \hfill
    \begin{minipage}[t]{0.48\textwidth}
        \centering
        \includegraphics[width=\linewidth]{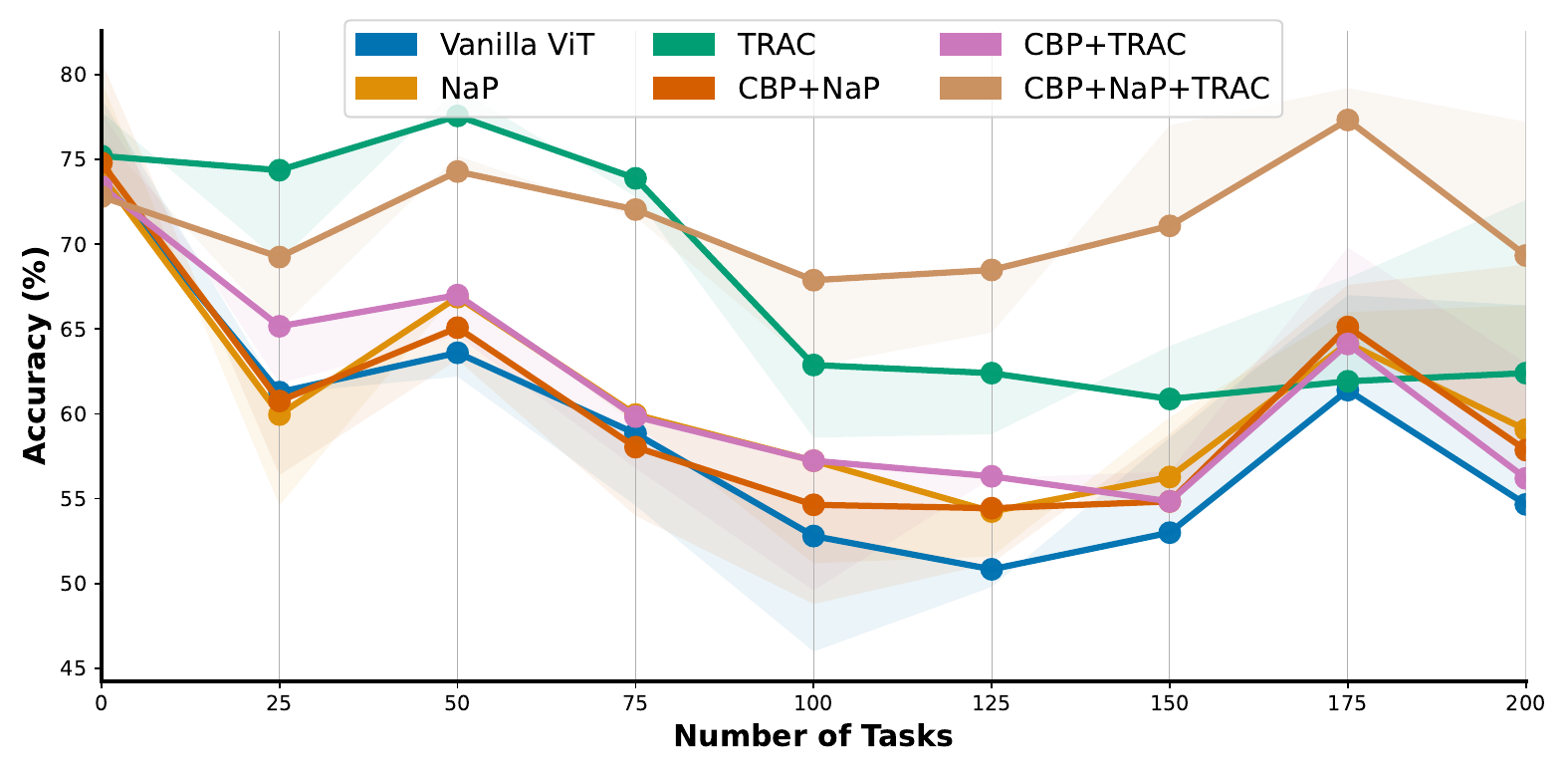}
        \subcaption{}
        \label{fig:all_accuracies_comparison_moving_average}
    \end{minipage}
    \vfill
    \begin{minipage}[t]{\linewidth}
        \centering
        \includegraphics[width=\linewidth]{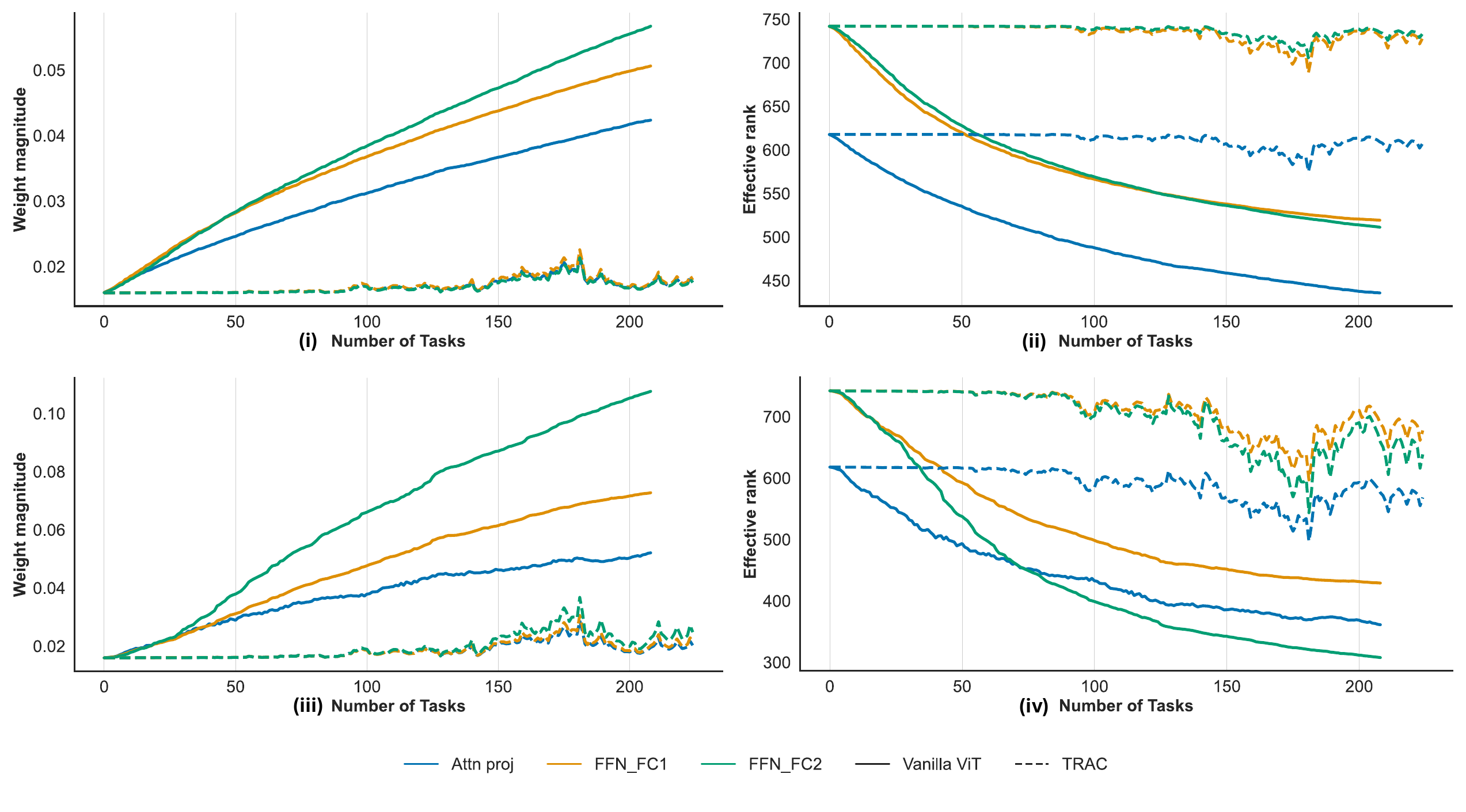}
        \subcaption{}
        \label{fig:fig_weight_metrics_vit_trac}
    \end{minipage}
    \caption{(a) Performance comparison between vanilla ViT and ViT with CBPs. For CBP1–CBP5, the maturity thresholds are 400, 500, 1200, 600, and 1200. The replacement rates are $8 \times 10^{-6}$, $1 \times 10^{-5}$, $3 \times 10^{-6}$, $5 \times 10^{-6}$, and $8 \times 10^{-7}$. CBP2 and CBP3 apply CBP to the first block, whereas CBP1, CBP4, and CBP5 apply it to all blocks. (b) Performance comparison between vanilla ViT and representative plasticity loss mitigation methods. (c) Weight magnitude of the 1st block (i) and 12th block (iii), and the corresponding effective rank of the 1st block (ii) and 12th block (iv).}
\end{figure}

\subsection{Evaluation of Plasticity Loss Mitigation Methods}
To identify effective remedies for plasticity loss in ViTs, we adapt and evaluate representative strategies that have been evaluated in homogeneous architectures. These baselines span four categories: (1) \textbf{Reset-based Intervention}: continual back-propagation (CBP) \cite{dohare2024loss}; (2) \textbf{Normalization}: normalization-and-projection (NaP) \cite{lyle2024normalization}; (3) \textbf{Activation Function}: CReLU \cite{abbas2023loss}; and (4) \textbf{Optimizer}: TRAC \cite{muppidi2024fast}.

The CBP algorithm maintains model plasticity by identifying and replacing mature neurons that contribute little to learning. In our implementation, we apply CBP specifically to the FFN modules that contain non-linear activations, with five configurations ranging from conservative to aggressive replacement. However, as illustrated in Fig.~\ref{fig:cbp_accuracies_comparison_moving_average}, CBP shows marginal improvements in ViTs. This indicates that the structural complexity of attention-based models makes simple neuron-wise replacement insufficient for recovering global representation capacity, as the inter-dependency between attention and FFN layers remains unaddressed.

As shown in Fig.~\ref{fig:all_accuracies_comparison_moving_average}, we further evaluate NaP and alternative activations CReLU (Appendix, exhibiting worse initial performance and incurring double computational cost), both of which show negligible benefits for ViT plasticity. In contrast, the \textbf{optimization-based} method TRAC delivers a significant performance improvement. TRAC dynamically regulates the update step size, thereby effectively suppressing the explosion of weight magnitudes (Fig.~\ref{fig:fig_weight_metrics_vit_trac}) and preserving a significantly higher effective rank across both shallow and deep blocks

\begin{finding}
    Empirical results suggest that strategies that regulate the optimization process tend to be more effective at preserving ViT plasticity than localized re-initialization or architectural modifications.
\end{finding}

\section{Adaptive Rank-Reshaping via Online Windowed Covariance}

\subsection{From Scaling Step Size to Correcting Direction}
While TRAC achieves strong empirical performance by dynamically rescaling the learning rate to counteract non-stationary gradients, such step-size control is inherently direction-agnostic. It is well established that first-order methods (\textit{e.g.}, SGD) update parameters along the instantaneous gradient direction, differing in update magnitude (In Eq.~\eqref{arrow1}, $P=I$). However, in continual learning, gradients tend to align with a limited set of dominant directions shaped by earlier tasks \cite{tang2025mitigating}. This directional bias constrains the model's ability to effectively incorporate information from new data. Plasticity loss is not solely a matter of step size, but fundamentally a geometric issue arising from gradient direction concentration \cite{castanyer2025stable}.
\begin{small}
\begin{equation}
\label{arrow1}
\Delta = -\eta P g
\end{equation}
\end{small}

Second-order optimization methods address plasticity loss by explicitly incorporating local curvature of the loss landscape. Consider the second-order Taylor expansion of the objective around $\theta$:
\begin{small}
\begin{equation}
\mathcal{L}(\theta + \Delta) \approx \mathcal{L}(\theta) + g^\top \Delta + \frac{1}{2}\Delta^\top H \Delta,
\end{equation}
\end{small}
where $g = \nabla_\theta \mathcal{L}(\theta)$ and $H = \nabla_\theta^2 \mathcal{L}(\theta)$ is the Hessian matrix. Minimizing this quadratic model yields the Newton update:
\begin{small}
\begin{equation}
\Delta_{\mathrm{Newton}} = - H^{-1} g.
\end{equation}
\end{small}

Since $H$ is a real symmetric matrix, it admits an eigen-decomposition. This update rescales gradient components according to the curvature: directions associated with large curvature (large eigenvalues of $H$) are attenuated, whereas directions with small curvature are amplified. This direction exploration is particularly relevant for plasticity because gradients become increasingly aligned with a limited set of dominant directions inherited by earlier tasks, making optimization low-dimensional and hindering adaptation.

In practice, exact Newton updates are rarely feasible because $H$ is expensive to compute and invert. A classical remedy is the Levenberg-Marquardt (LM) method \cite{more2006levenberg}, or using the Fisher Information Matrix $F$ to replace $H$ \cite{wang2025avoid}.

\subsection{The ARROW Implementation}
Motivated by the previous findings, we propose ARROW, a geometry-aware optimizer that approximates second-order behavior by reshaping gradient directions using a low-rank, online curvature proxy. At each step $t$, ARROW computes:
\begin{small}
\begin{equation}
\Delta \theta_t = - \eta_t \, (\alpha_t I + \beta C_t)^{-1} g_t.
\end{equation}
\end{small}
where $g_t$ is the current gradient, $C_t$ is a windowed gradient covariance estimate, $\alpha>0$ is a damping factor, $\beta \ge0$ controls curvature strength, and $\eta_t$ is the step size.

For the curvature proxy $C_t$, it is defined as:
\begin{small}
\begin{equation}
C_t = \frac{1}{W} \sum_{i=t-W+1}^{t} g_i g_i^\top .
\end{equation}
\end{small}
where $W$ is a fixed window size. This construction yields a positive semi-definite matrix with rank at most $W$, capturing the dominant update directions induced by recent data to adapt to non-stationary gradient distributions.

ARROW can be understood through the eigendecomposition of $C_t$. In \cref{eq8}, along an eigen-vector $u_i$ with eigenvalue $\lambda_i$, the update is rescaled by: $1/(\alpha_t + \beta\lambda_t)$. Consequently, directions with large $\lambda_i$ corresponding to frequently activated high-curvature subspaces are suppressed. On the other hand, directions with small $\lambda_i$ are relatively amplified. This reweighting across a broader set of directions effectively expanding the update subspace. In continual learning, this mechanism directly counteracts the collapse of effective rank \cite{chaudhry2020continual}.
\begin{small}
\begin{equation}\label{eq8}
(\alpha_t I + \beta C_t)^{-1} u_i
= \frac{1}{\alpha_t + \beta \lambda_i}\, u_i .
\end{equation}
\end{small}

ARROW remains efficient by exploiting the low-rank structure of $C_t$. Using the Woodbury identity, the inverse $(\alpha_t I + \beta C_t)^{-1}$ can be regarded as a $W^2$ system. 
\begin{small}
\begin{equation}
G_t = [g_{t-W+1}, \dots, g_t] \in \mathbb{R}^{d \times W},
U_t = \sqrt{\beta / W}\, G_t .
\end{equation}
\end{small}
\begin{small}
\begin{equation}
\begin{aligned}
(\alpha_t I + U_t U_t^\top)^{-1} g_t
&= \frac{1}{\alpha_t} g_t - \frac{1}{\alpha_t^2}
U_t\left(I + \frac{1}{\alpha_t} U_t^\top U_t\right)^{-1}
U_t^\top g_t .
\end{aligned}
\end{equation}
\end{small}

In practice, we use RMS-like or stochastic gradient descent (SGD) to warm up the updating before the window is completely filled. For $\alpha$, using an exponential descent strategy is possibly feasible.

\section{Experiments}
In this section, we conduct experiments to investigate the following research questions:
\begin{itemize}
    \item \textbf{Q1}: What are the most suitable configurations of $\alpha$, $\beta$, window size $W$, and the warm-up strategy?
    \item \textbf{Q2}: How effective are the estimated second-order terms in ARROW, and which blocks are most affected? Do these effects align with the findings drawn in \cref{sec:loss}?
    \item \textbf{Q3}: Does ARROW outperform competitive methods for maintaining plasticity across different categories?
    \item \textbf{Q4}: What are the computational costs of ARROW in terms of time and memory overhead?
\end{itemize} 

\subsection{Experimental Setup}
\subsubsection{Benchmark Selection}
We evaluate the methods on CIFAR100 and ImageNet-R \cite{hendrycks2021many}, two widely used benchmarks for continual learning. CIFAR-100 consists of 100 classes with 60,000 $32 \times 32$ images with low resolution. ImageNet-R contains 200 object classes, comprising 30,000 images rendered in diverse artistic styles (\textit{e.g.}, sketches, cartoons). The stylistic variations introduce substantial distribution shifts. The employment of two datasets makes it a challenging benchmark for evaluating robustness and plasticity under non-stationary data.

For the task-incremental continual learning scenario, we construct disjoint task streams by partitioning classes. For CIFAR-100, we construct task streams with 10, 20, and 25 tasks, corresponding to 10, 5, and 4 classes per task, respectively. For ImageNet-R, task streams are length of 20, 40, and 50. This design allows us to examine plasticity ranging from moderate to highly challenging continual learning settings.

\subsubsection{Algorithmic Baselines}
We compare proposed ARROW against L2P \cite{wang2022learning}, TRAC \cite{muppidi2024fast}, NaP \cite{lyle2024normalization}, and CBP \cite{dohare2024loss}, spanning architectural, optimization, normalization, and re-initialization baselines for plasticity loss mitigation.
\subsubsection{Evaluation Metrics}
The main metrics to evaluate the loss of plasticity is Average Accuracy across Tasks (AAT). The accuracy of classification directly reflect the agent's capability to acquire new knowledge.
\subsubsection{Specific Settings}
All experiments are conducted on a cluster with 10 NVIDIA RTX 4090 GPUs. Models are trained on a single GPU without model parallelism. Singular Value Decomposition (SVD) for rank-based metrics is computed during evaluation only and does not significantly affect training efficiency.

\subsection{Results Analysis}
\subsubsection{Hyperparameter Adjustment}
ARROW introduces hyperparameters, including the damping factor $\alpha$, the curvature strength $\beta$, the warm-up mode (SGD or RMS-like), and the window size $W$.

$\alpha$ is the most influential hyperparameter, while $\beta$ modulates the strength of curvature correction. The window size $W$ controls the amount of historical gradient information and affects computational cost. We evaluate $\alpha$ in $[1, 10^{-5}]$ and $\beta$ in $[0.3, 1.3]$, with window sizes between 10 and 30 and two warm-up strategies: SGD and RMS-like.
\begin{table*}[t]
\centering
\caption{Hyperparameter study of ARROW. Left: Performance across different $\alpha$, $\beta$, window sizes $W$ and warm-up modes. Right: Fine-grained tuning with fixed $W=20$ and RMS-like warm-up on CIFAR100 (10 tasks).}
\label{tab:arrow_hyper}

\begin{minipage}{0.53\linewidth}
\centering
\tiny
\begin{tabular}{c c c c}
\toprule
$\alpha$ & $W=10$ & $W=20$ & $W=30$ \\
\midrule
1       & -- & 59.51 ($\beta$=1, SGD) & -- \\
0.5     & -- & 58.08 ($\beta$=1, SGD) & 59.51 ($\beta$=1, SGD) \\
0.05    & -- & 60.45 ($\beta$=1, SGD) & -- \\
0.01    & -- & 60.51 ($\beta$=1, RMS) & -- \\
0.003   & -- & 60.51 ($\beta$=1, RMS) & -- \\
0.001   & 58.50 ($\beta$=1, SGD) & 60.41 ($\beta$=0.9, RMS) & 58.75 ($\beta$=1, SGD) \\
0.0001  & -- & 60.17 ($\beta$=1, SGD) & -- \\
\bottomrule
\end{tabular}
\end{minipage}
\hfill
\begin{minipage}{0.35\linewidth}
\centering
\scriptsize
\begin{tabular}{c c c c c}
\toprule
Config & $\alpha$ & $\beta$ & RMS & Avg Acc (\%) \\
\midrule
C1 & $1\mathrm{e}{-2}$ & 1.0 & \checkmark & 63.17 \\
C3 & $1\mathrm{e}{-3}$ & 0.9 & \checkmark & \textbf{63.96} \\
C4 & $3\mathrm{e}{-3}$ & 1.0 & \checkmark & 63.18 \\
C5 & $5\mathrm{e}{-2}$ & 1.0 & \checkmark & 61.46 \\
\bottomrule
\end{tabular}
\end{minipage}

\end{table*}
We first perform a coarse search to identify a promising region for $\alpha$, $\beta$, and the warm-up mode. As shown in \cref{tab:arrow_hyper} (left), the results indicate that $\alpha$ values in the range $5 \times 10^{-2}$ to $1 \times 10^{-3}$ consistently yield better performance. Based on this analysis, we fix the warm-up strategy to RMS-like updates and window size $W=20$ for subsequent experiments. The final hyperparameter configuration of ARROW is summarized in \cref{tab:arrow_hyper} (right). We additionally evaluate exponential decay schedules for $\alpha$ with rates $0.999$, $0.99$, $0.98$, and $0.97$, considering both global and per-task decay schemes (Appendix). However, neither strategy improves over the best fixed-parameter configuration.


\subsubsection{Ablation Study}
To validate the contribution of individual components in ARROW, we disable key terms and vary the application scope across blocks. Specifically, we test settings with $\alpha = 1$, $\beta = 0$, and apply ARROW to different subsets of attention blocks (all blocks, the last six blocks, and the last block). When $\beta = 0$ and $\alpha = 1$, ARROW degenerates into a first-order optimizer, removing both directional damping and curvature-based reshaping.

As shown in \cref{ablation}, applying ARROW to the last attention blocks yields the best performance. This confirms the necessity of its key components and supports the findings in \cref{sec:loss} that deep attention modules are the contributors to plasticity degradation and the effective targets for geometry-aware optimization.
\begin{table}[t]
\centering
\scriptsize
\caption{Ablation study results of ARROW with controlling terms of $\alpha, \beta$, and effected blocks.}
\label{ablation}
\renewcommand{\arraystretch}{0.9}
\begin{tabular}{l c c c c}
\toprule
Method & Location & $\alpha$ & $\beta$ & AAT (\%) \\
\midrule
ARROW     & Last 1    & $1.0 \times 10^{-3}$ & 0.9 & \textbf{64.21} \\
ARROW\_b0 & ALL    & $1.0 \times 10^{-3}$ & 0   & 59.85 \\
ARROW\_a1 & ALL    & $1.0$                & 0.9 & 59.31 \\
ARROW\_6  & Last 6 & $1.0 \times 10^{-3}$ & 0.9 & 63.94 \\
ARROW\_1  & ALL & $1.0 \times 10^{-3}$ & 0.9 & 63.96 \\
\bottomrule
\end{tabular}
\end{table}

\begin{table}[t]
\centering
\scriptsize
\caption{Comparison of plasticity performance (AAT \%) on CIFAR100 and ImageNet-R across different length of task streams.}
\label{comparison}
\begin{tabular}{l c c c c c c}
\toprule
& \multicolumn{3}{c}{CIFAR100} & \multicolumn{3}{c}{ImageNet-R} \\
\cmidrule(lr){2-4}\cmidrule(lr){5-7}
Method & 10 Task & 20 Task & 25 Task & 20 Task & 40 Task & 50 Task \\
\midrule
Baseline & 57.21 & 66.53 & 70.93 & 28.16 & 36.94 & 40.61 \\
CBP      & 57.43 & 66.18 & 67.09 & 27.27 & 36.35 & 41.96 \\
NaP      & 58.62 & 66.93 & 71.46 & 28.20 & 37.35 & 42.06 \\
L2P     & 60.17 & 67.54& 71.13 & 30.83 & 35.06 & 40.88 \\
TRAC     & 63.58 & 73.32 & 72.19 & 27.98 & 37.17 & 41.05 \\
ARROW    & \textbf{64.21} & \textbf{73.52} & \textbf{73.89} & \textbf{37.20} & \textbf{37.59} & \textbf{43.40} \\
\bottomrule
\end{tabular}
\end{table}
\subsubsection{Performance Comparison}
As shown in \cref{comparison}, ARROW achieves the highest AAT across all task streams and datasets. It slightly outperforms TRAC and L2P on most streams while consistently surpassing the remaining baselines. These results confirm the effectiveness of reshaping update directions via covariance-based geometry awareness, particularly in later stages of continual learning where plasticity degradation is most severe. Additionally, ARROW ($W=20$) has training time and GPU memory usage comparable to the vanilla ViT.
\section{Conclusion}
We present a systematic study of plasticity loss in Vision Transformers under task-incremental continual learning. Our analysis shows that plasticity degradation is strongly depth-dependent, with deep attention modules and FFNs exhibiting pronounced instability under continual distribution shifts. Evaluating representative mitigation strategies further reveals that optimization-based approaches are more effective than re-initialization or architectural modifications. Motivated by these findings, we propose \textbf{ARROW}, a geometry-aware optimizer that approximates second-order behavior by reshaping gradient directions using an online low-rank curvature proxy. Experiments show that ARROW achieves superior overall performance.

Nevertheless, ARROW introduces additional hyperparameters for curvature damping and windowed covariance estimation, making the method sensitive to tuning. Moreover, when models have different base performance levels, AAT may not always serve as a fully objective metric for evaluating improvements in plasticity.
\bibliographystyle{unsrt}  
\bibliography{main}
\newpage
\appendix
\section{Appendix}
\subsection{Feature Metrics of ViT}
\label{app:feature}
The feature metrics of 12 blocks in ViT. Each block involves embedding layers, normalization 1, attention output, residual structure 1, FFN's FC layer 1, FFN's FC layer 2, and the residual structure 2.
\begin{figure}[htbp]
  \centering
  \includegraphics[width=0.45\textwidth]{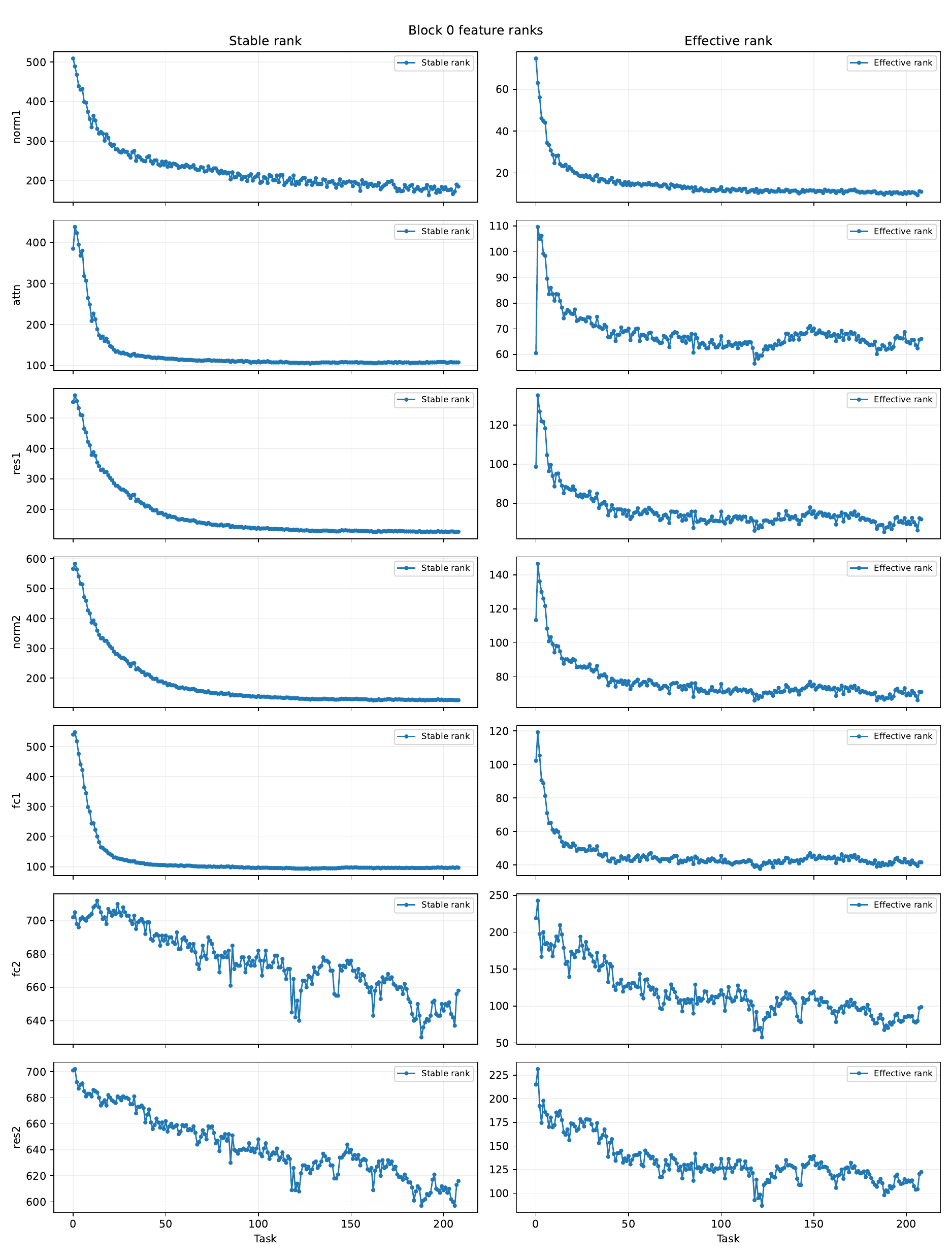}
  \includegraphics[width=0.45\textwidth]{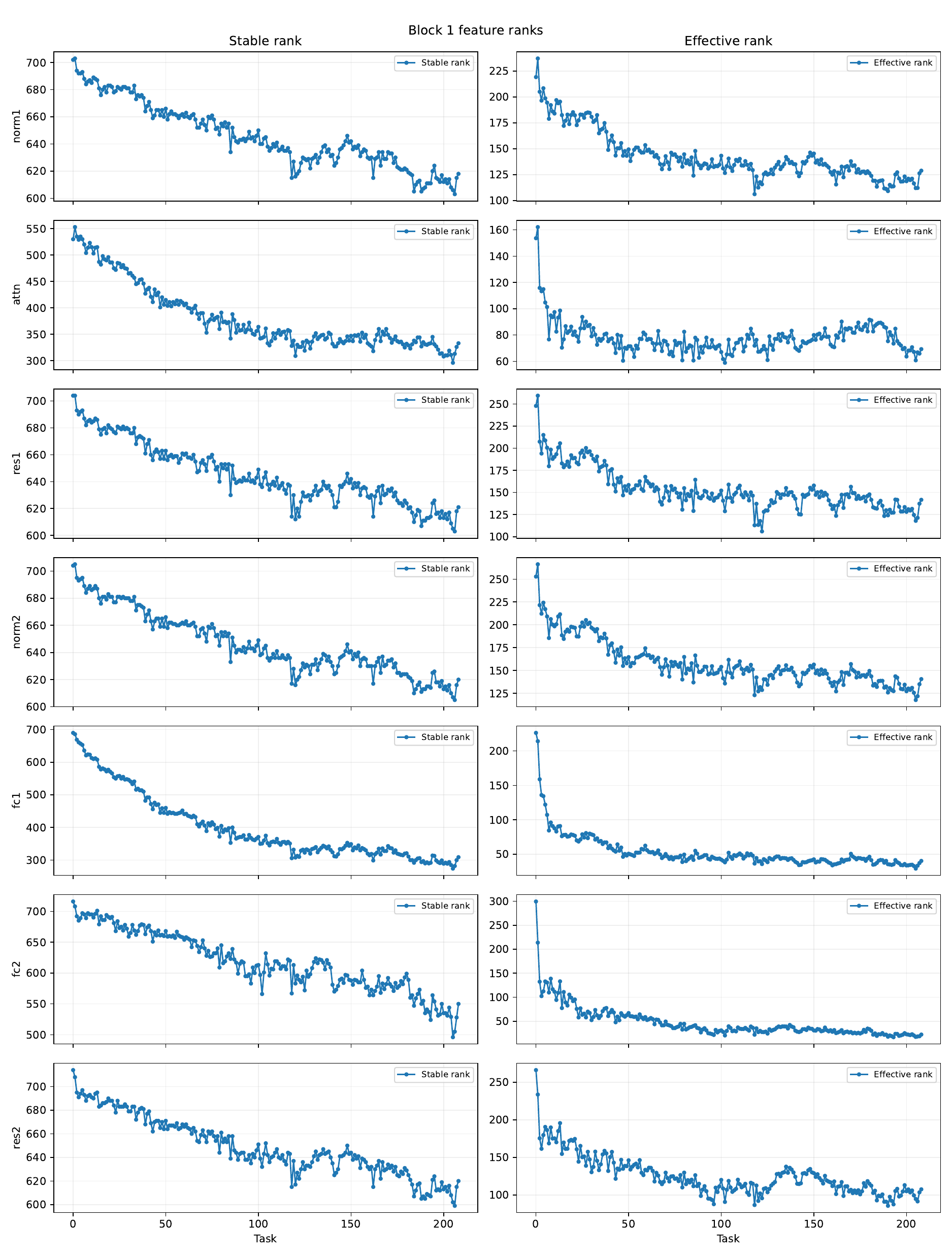}
  \caption{Stable rank and effective rank on first block (column left) and second block (column right) of ViT. The rows of sub-figures correspond with normalization 1, attention output, residual structure 1, normalization 2, FFN's FC layer 1, FFN's FC layer 2, and the residual structure 2 of vanilla ViT.}
\end{figure}
\begin{figure}[htbp]
  \centering
  \includegraphics[width=0.45\textwidth]{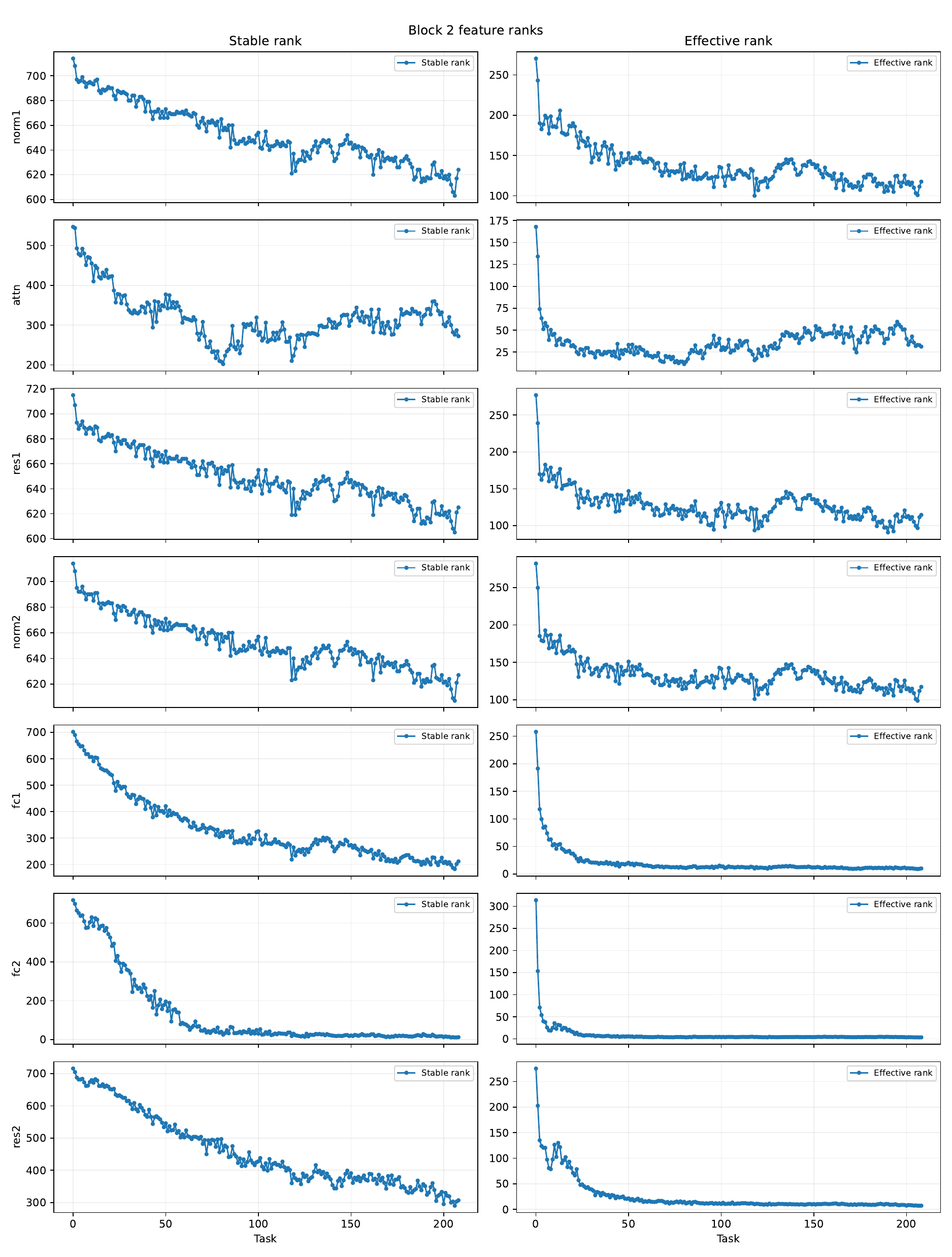}
  \includegraphics[width=0.45\textwidth]{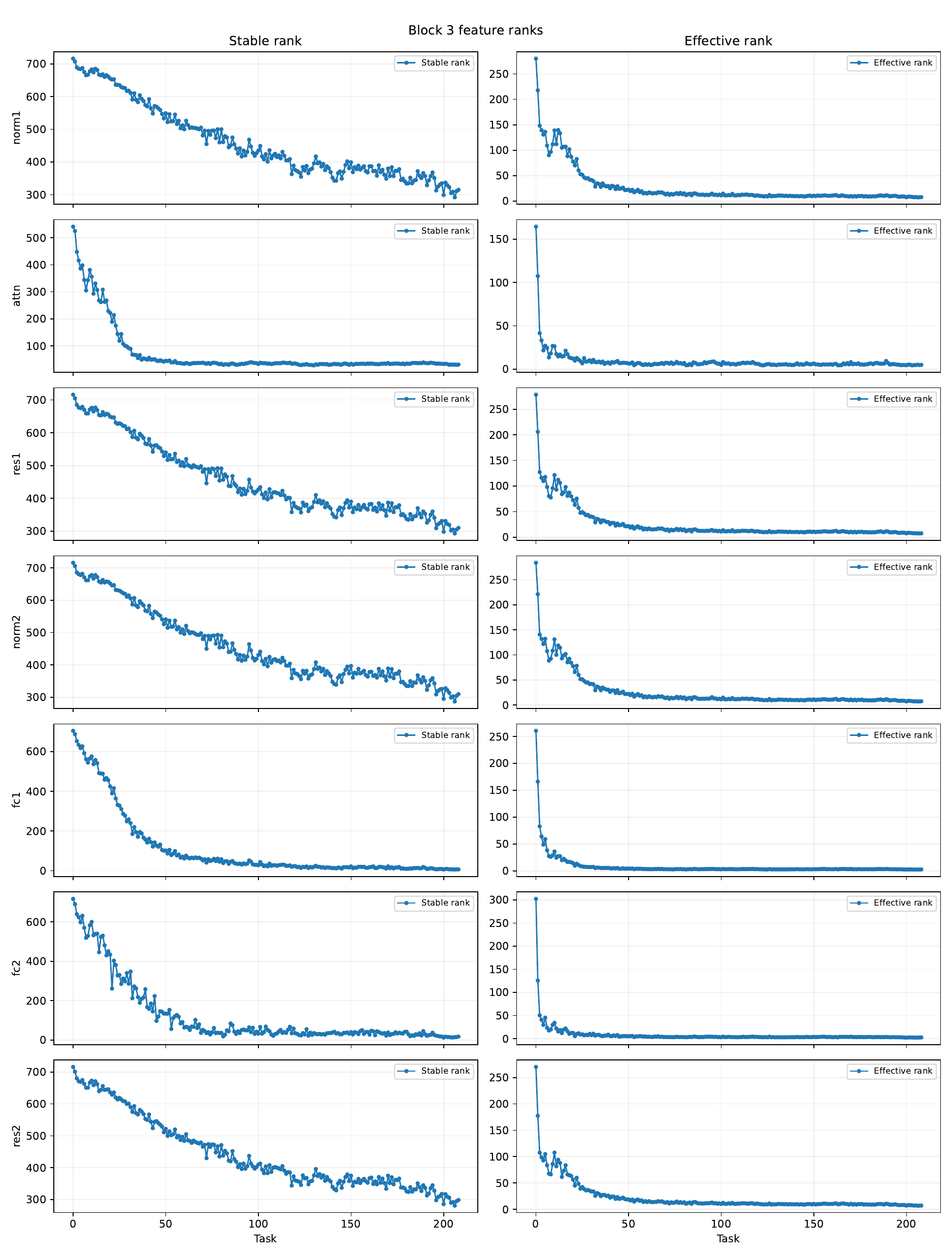}
  \caption{Stable rank and effective rank on 3rd block (column left) and 4th block (column right) of ViT. The rows of sub-figures correspond with normalization 1, attention output, residual structure 1, normalization 2, FFN's FC layer 1, FFN's FC layer 2, and the residual structure 2 of vanilla ViT.}
\end{figure}
\begin{figure}[htbp]
  \centering
  \includegraphics[width=0.45\textwidth]{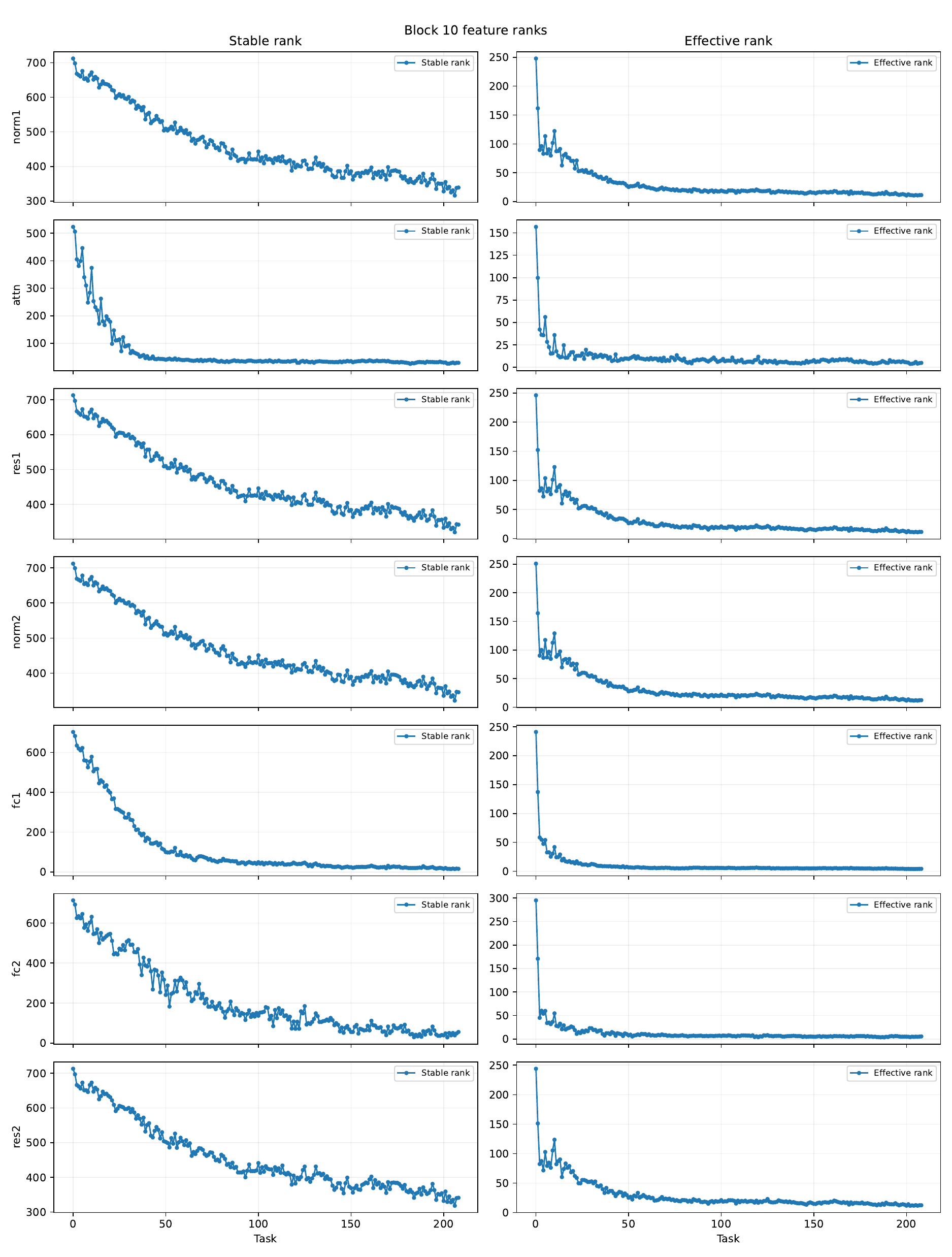}
  \includegraphics[width=0.45\textwidth]{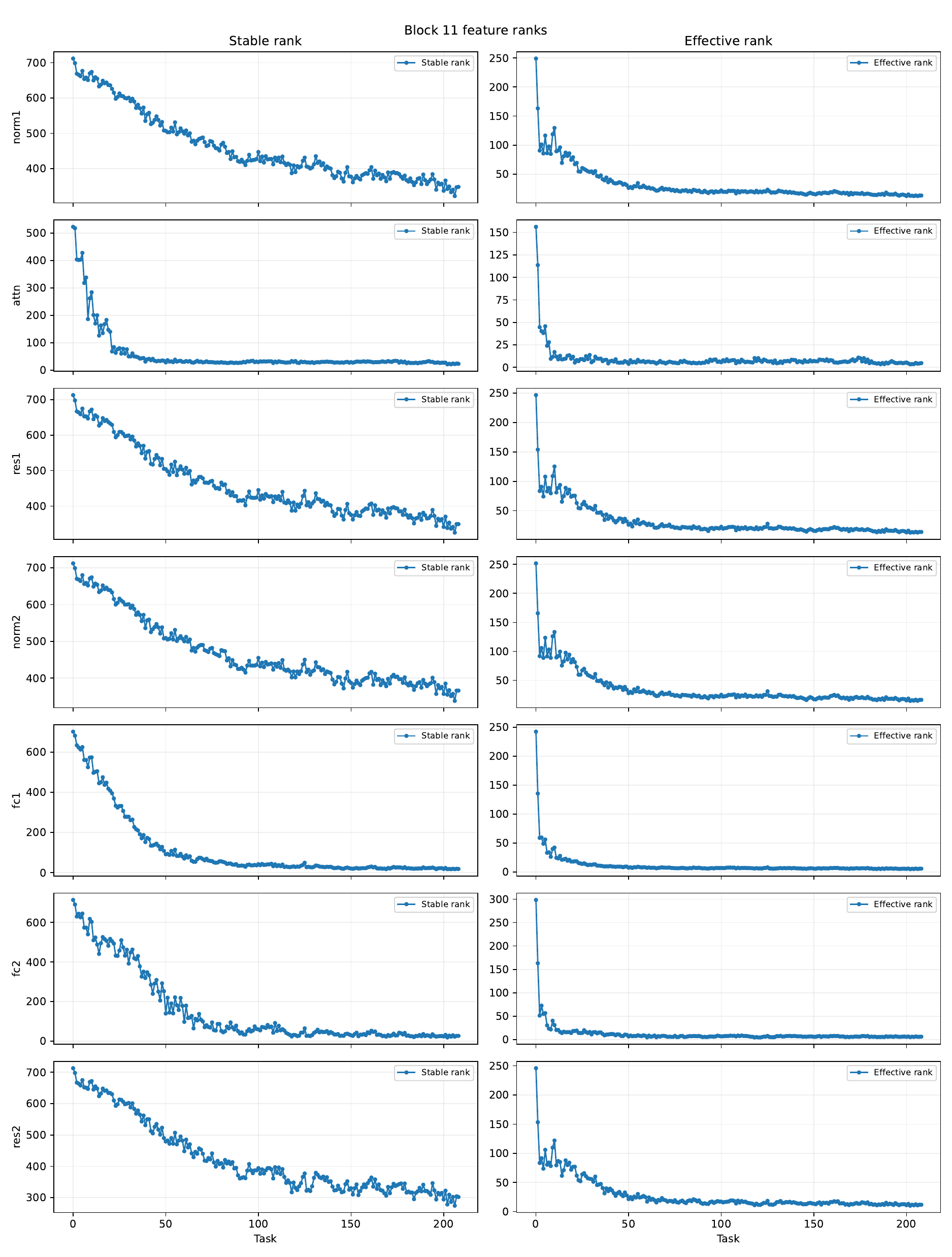}
  \caption{Stable rank and effective rank on 11th block (column left) and 12th block (column right) of ViT. The rows of sub-figures correspond with normalization 1, attention output, residual structure 1, normalization 2, FFN's FC layer 1, FFN's FC layer 2, and the residual structure 2 of vanilla ViT.}
\end{figure}
\begin{figure}[htbp]
  \centering
  \includegraphics[width=0.45\textwidth]{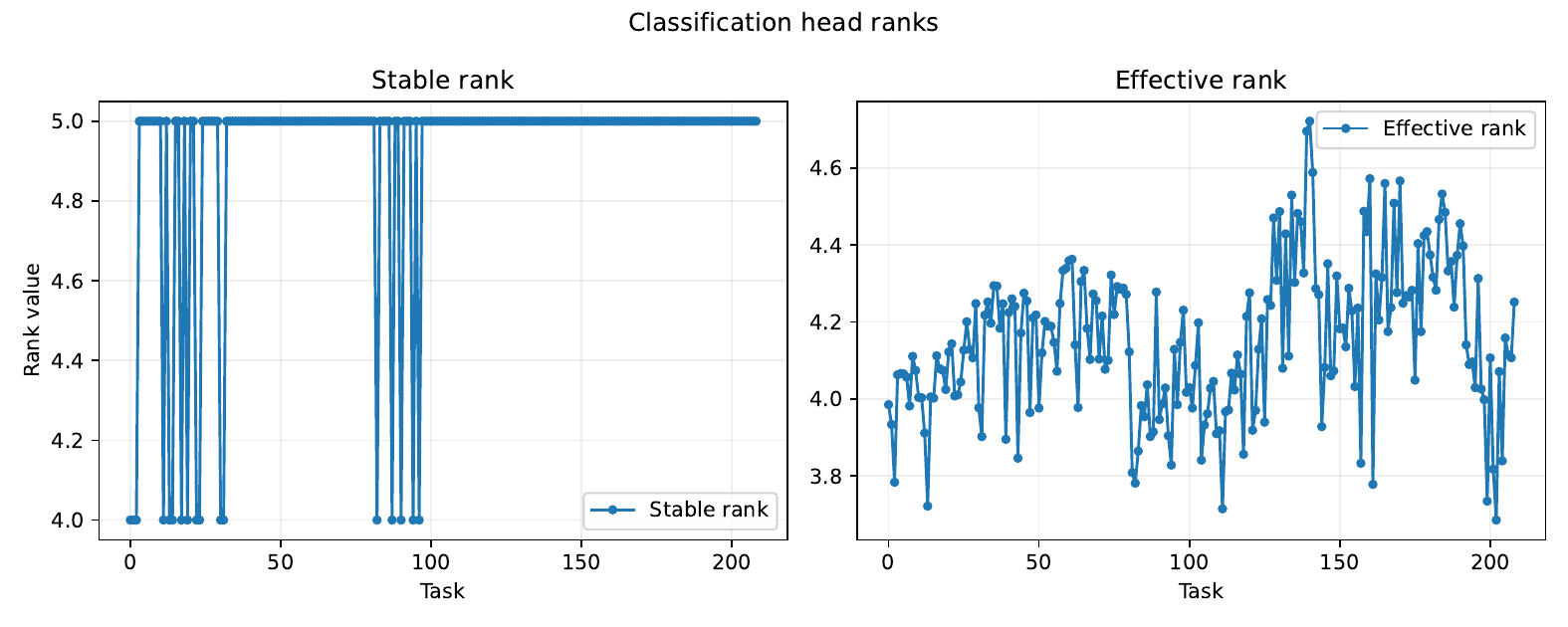}
  \includegraphics[width=0.45\textwidth]{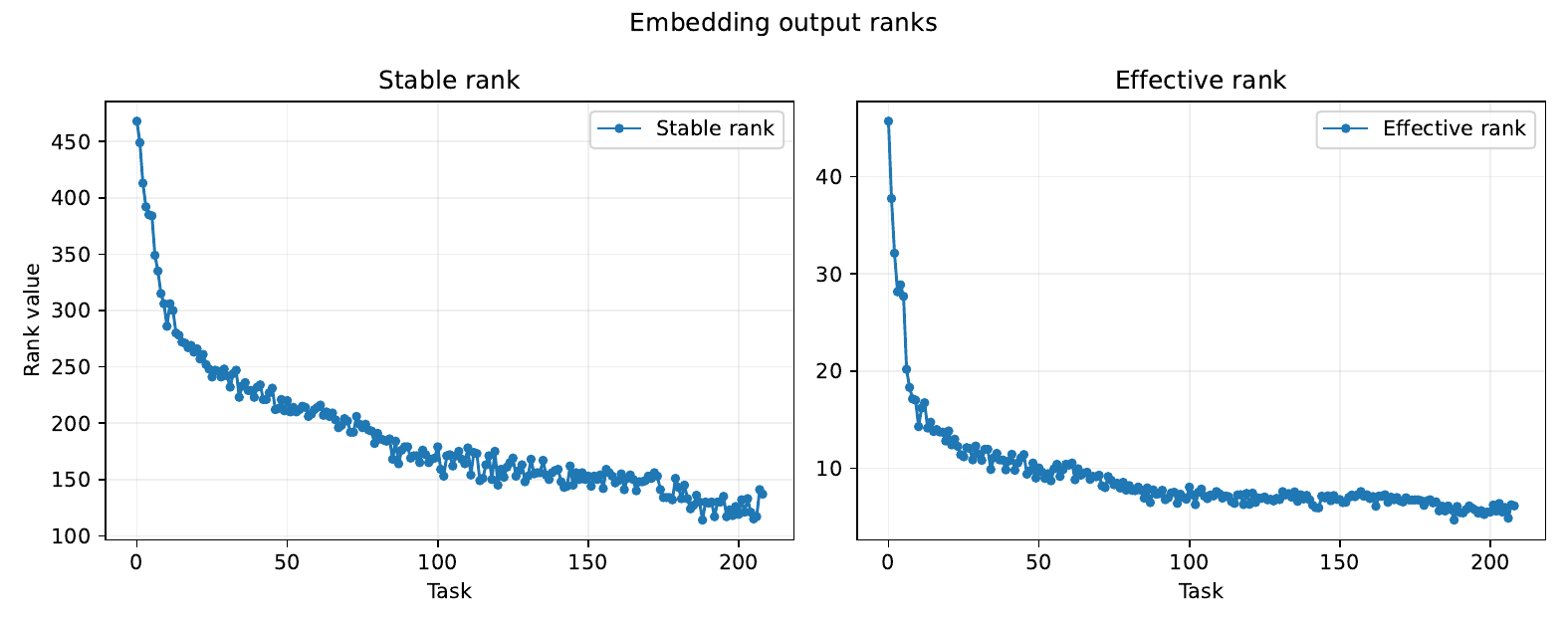}
  \caption{Stable rank and effective rank on embedding layer (column left) and classification head (column right) of vanilla ViT.}
\end{figure}
\newpage
\subsection{Weight Metrics of ViT}
\label{app:weight1}
The weight metrics of components in blocks of ViT, including effective rank and magnitude of 12 head, the FC layer 1, FC layer 2 of FFN, and attention projection.
\begin{figure}[htbp]
  \centering
  \includegraphics[width=0.45\textwidth]{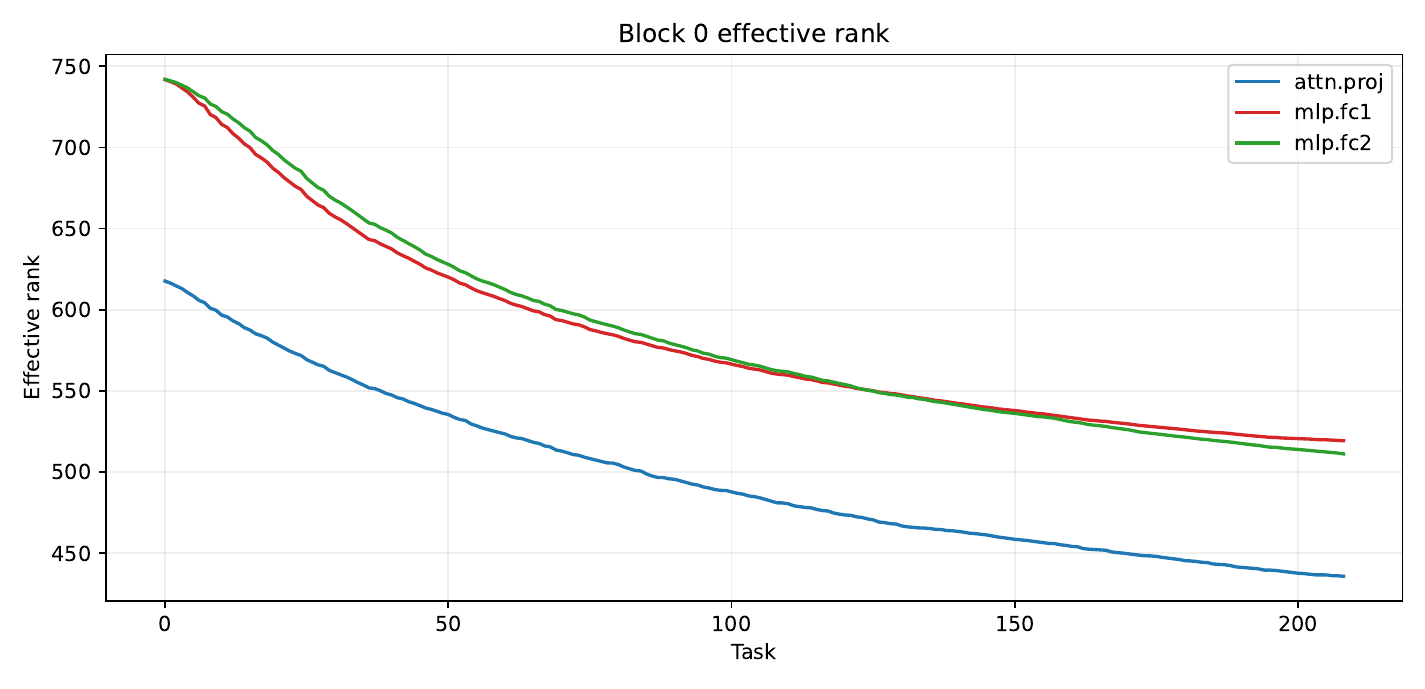}
  \includegraphics[width=0.45\textwidth]{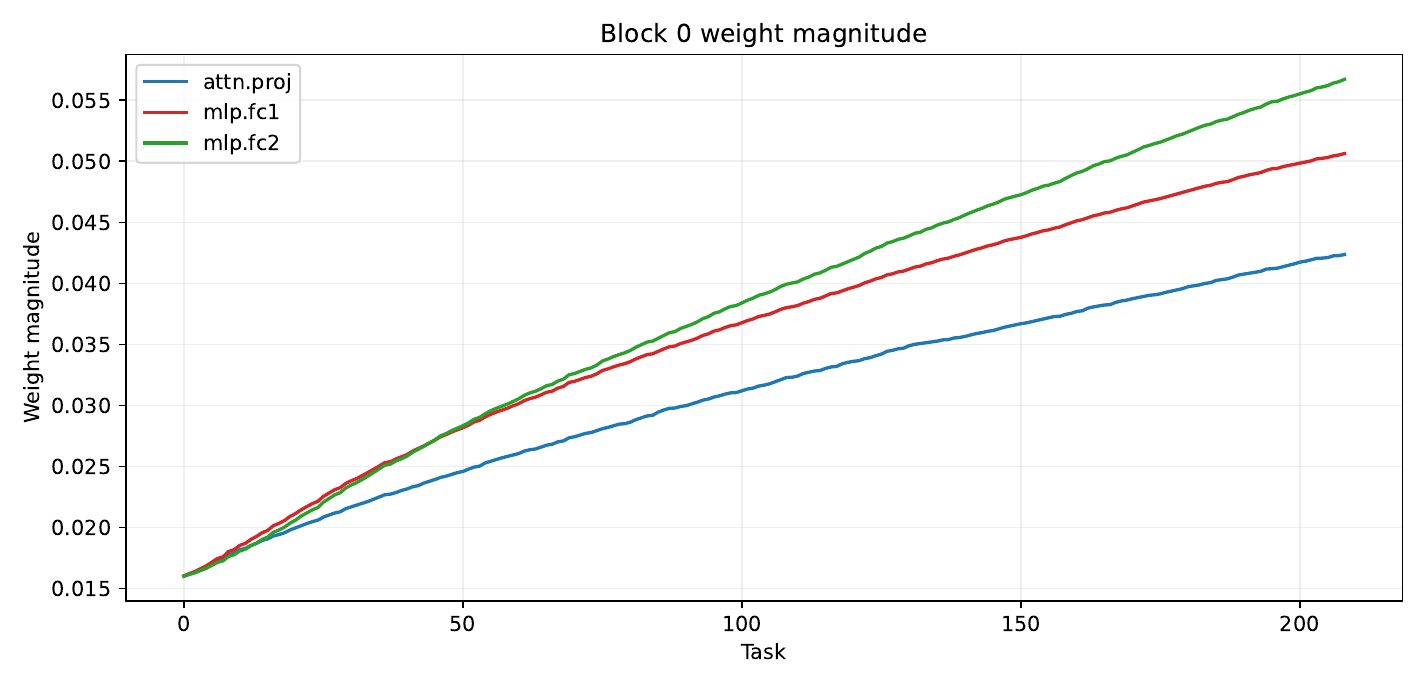}
  \caption{Effective rank (left) and weight magnitude (right) of attention projections and FFN of ViT, block 1. The blue curves represent the metrics of attention projection. The red and green curves represent the first and second FC layers in FFN.}
\end{figure}
\begin{figure}[htbp]
  \centering
  \includegraphics[width=0.45\textwidth]{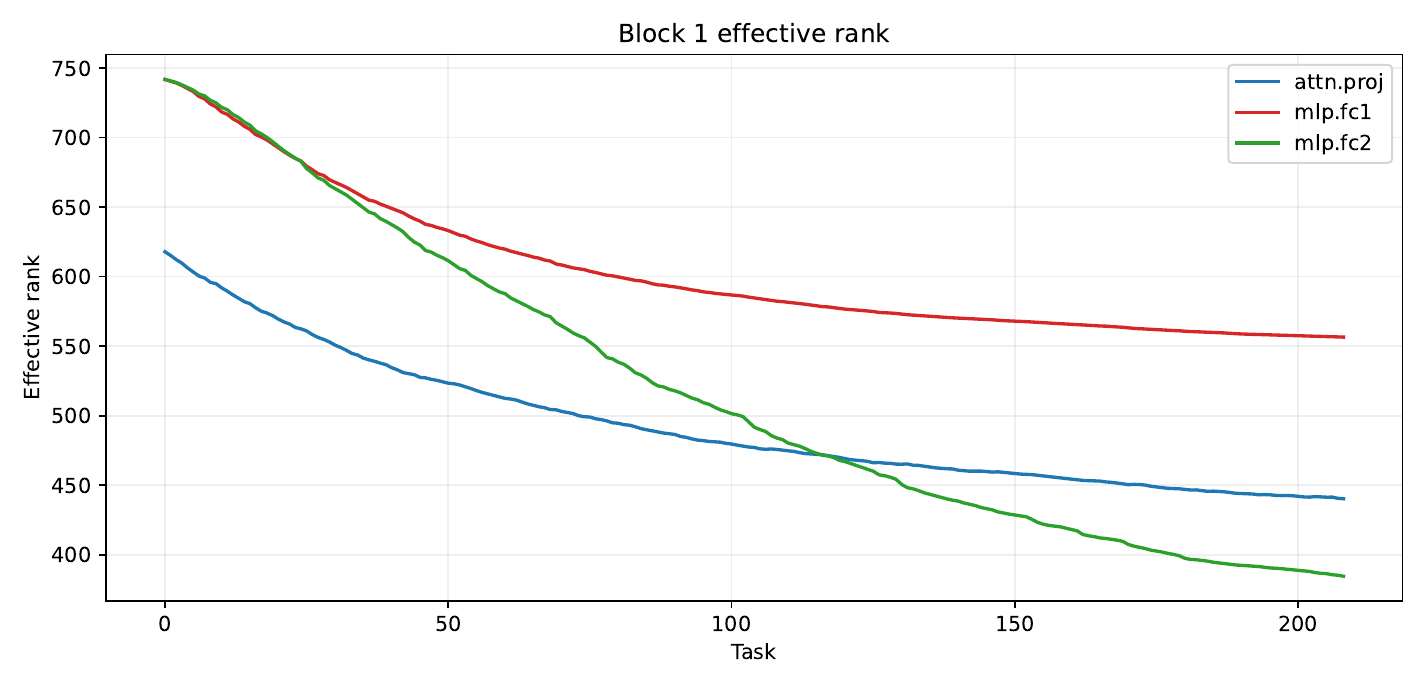}
  \includegraphics[width=0.45\textwidth]{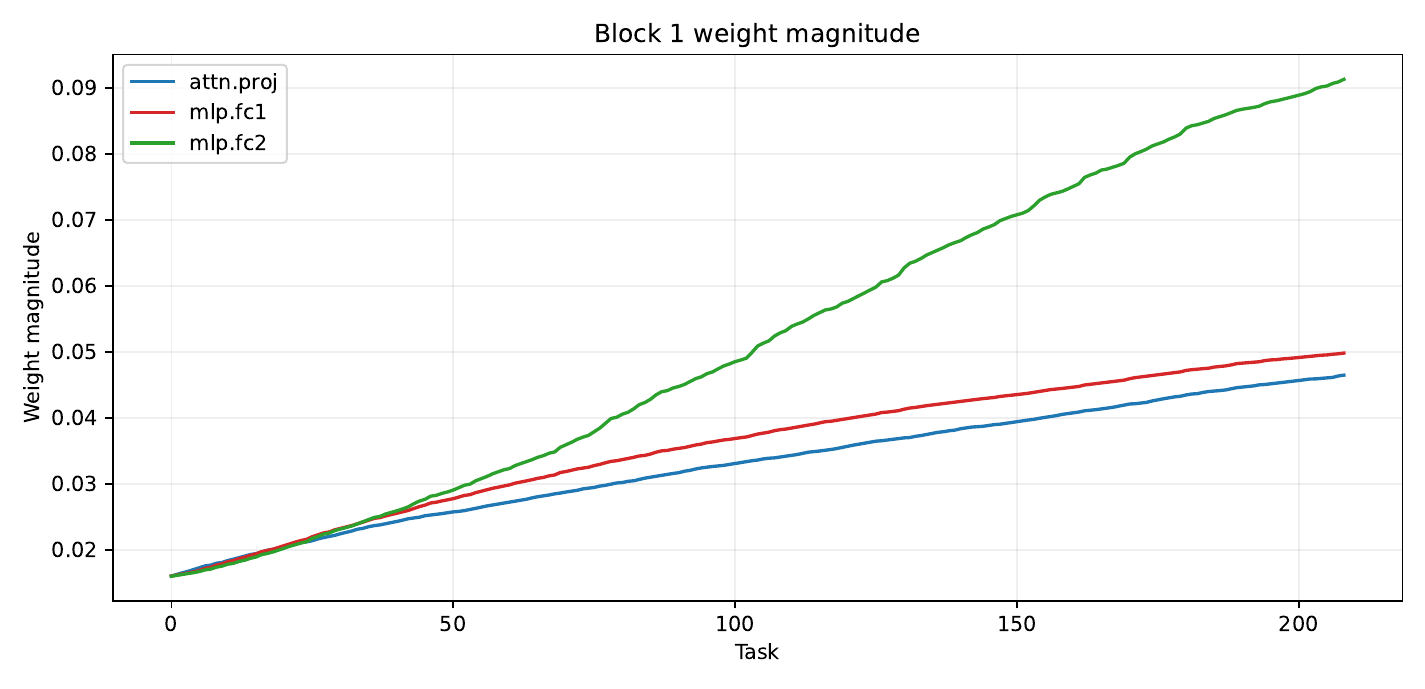}
  \caption{Effective rank (left) and weight magnitude (right) of attention projections and FFN of ViT, block 2. The blue curves represent the metrics of attention projection. The red and green curves represent the first and second FC layers in FFN.}
\end{figure}
\begin{figure}[htbp]
  \centering
  \includegraphics[width=0.45\textwidth]{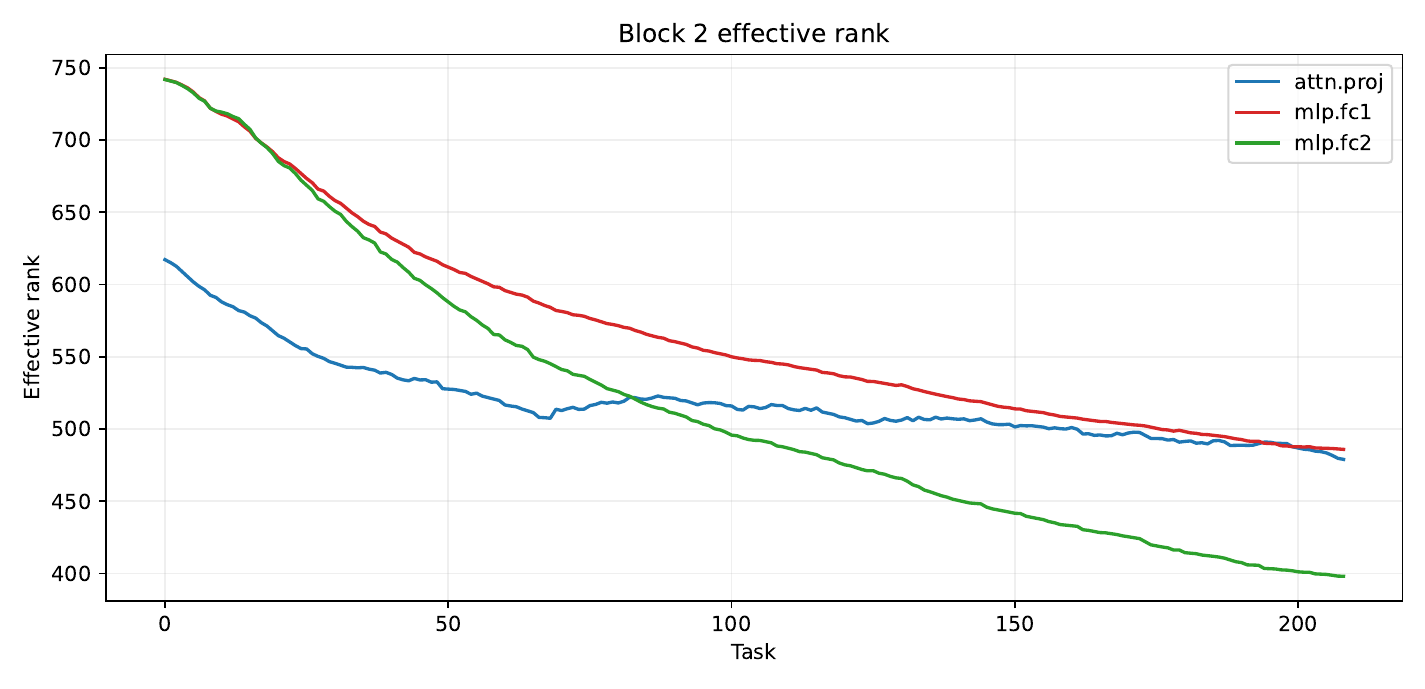}
  \includegraphics[width=0.45\textwidth]{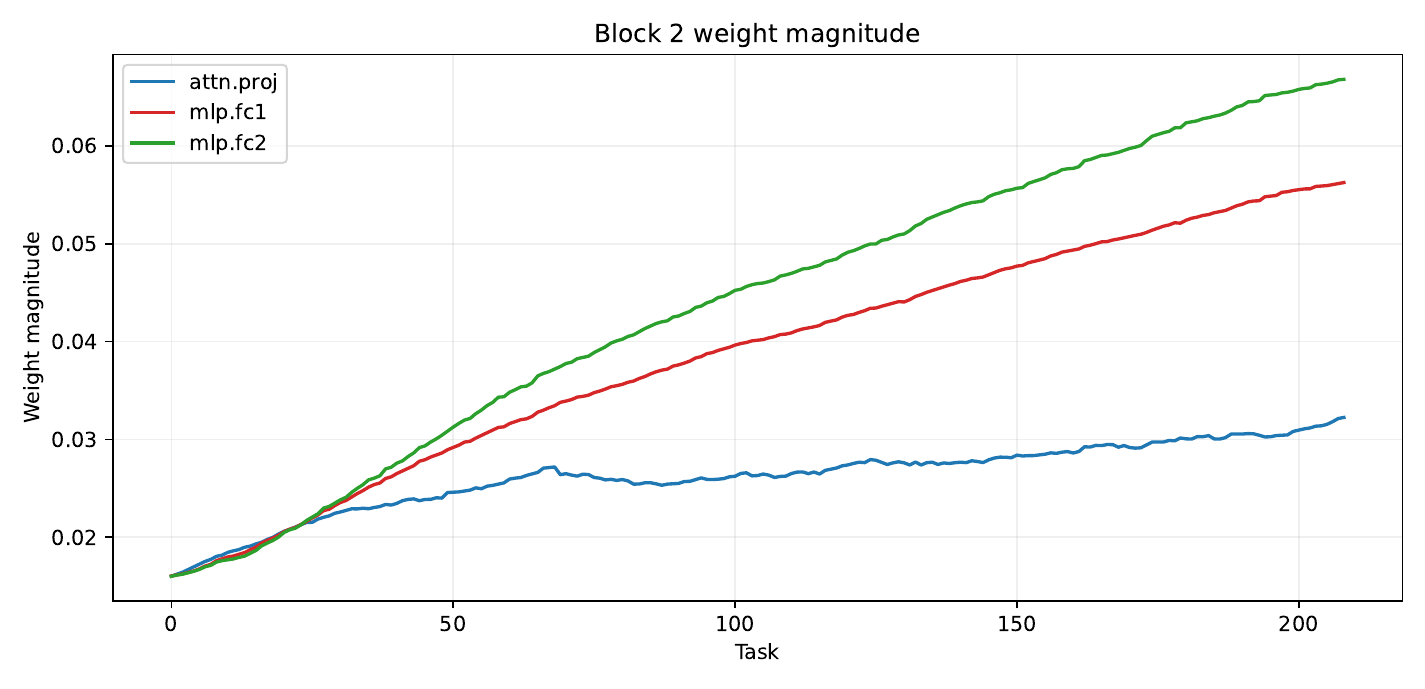}
  \caption{Effective rank (left) and weight magnitude (right) of attention projections and FFN of ViT, block 3. The blue curves represent the metrics of attention projection. The red and green curves represent the first and second FC layers in FFN.}
\end{figure}
\begin{figure}[htbp]
  \centering
  \includegraphics[width=0.45\textwidth]{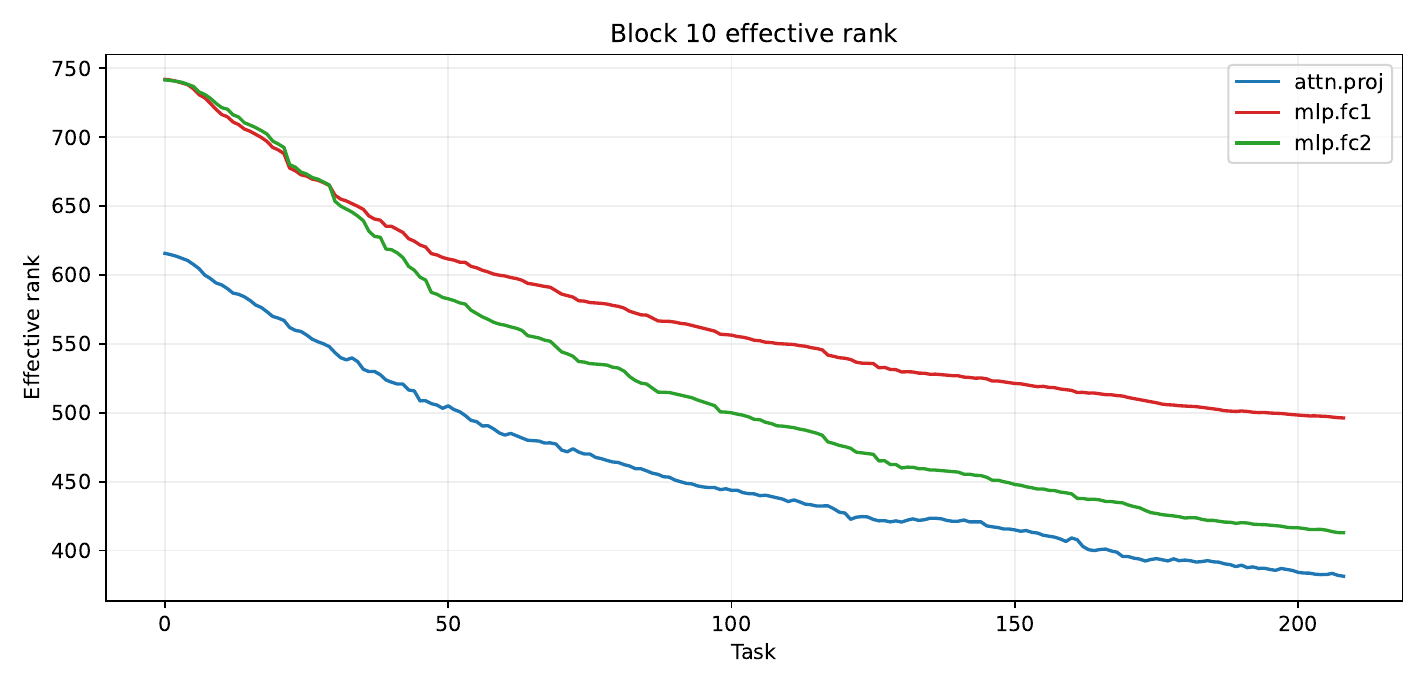}
  \includegraphics[width=0.45\textwidth]{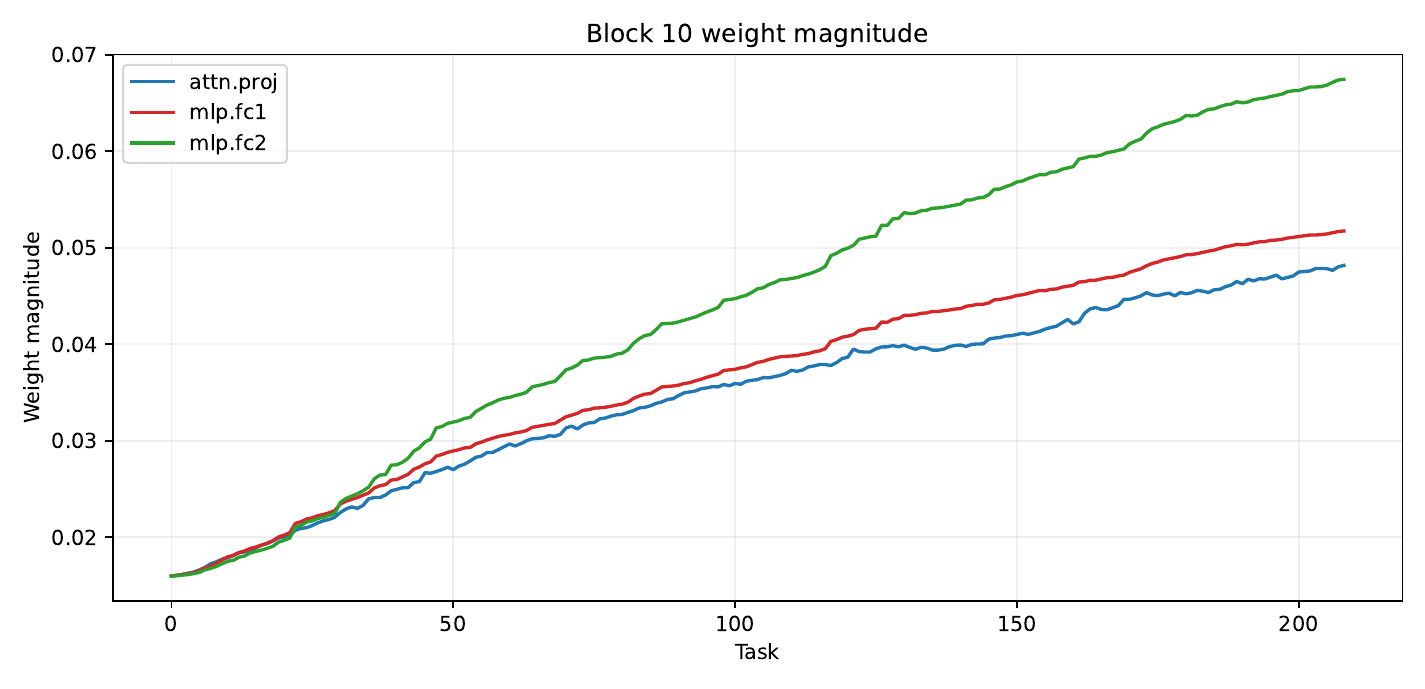}
  \caption{Effective rank (left) and weight magnitude (right) of attention projections and FFN of ViT, block 11. The blue curves represent the metrics of attention projection. The red and green curves represent the first and second FC layers in FFN.}
\end{figure}
\begin{figure}[htbp]
  \centering
  \includegraphics[width=0.45\textwidth]{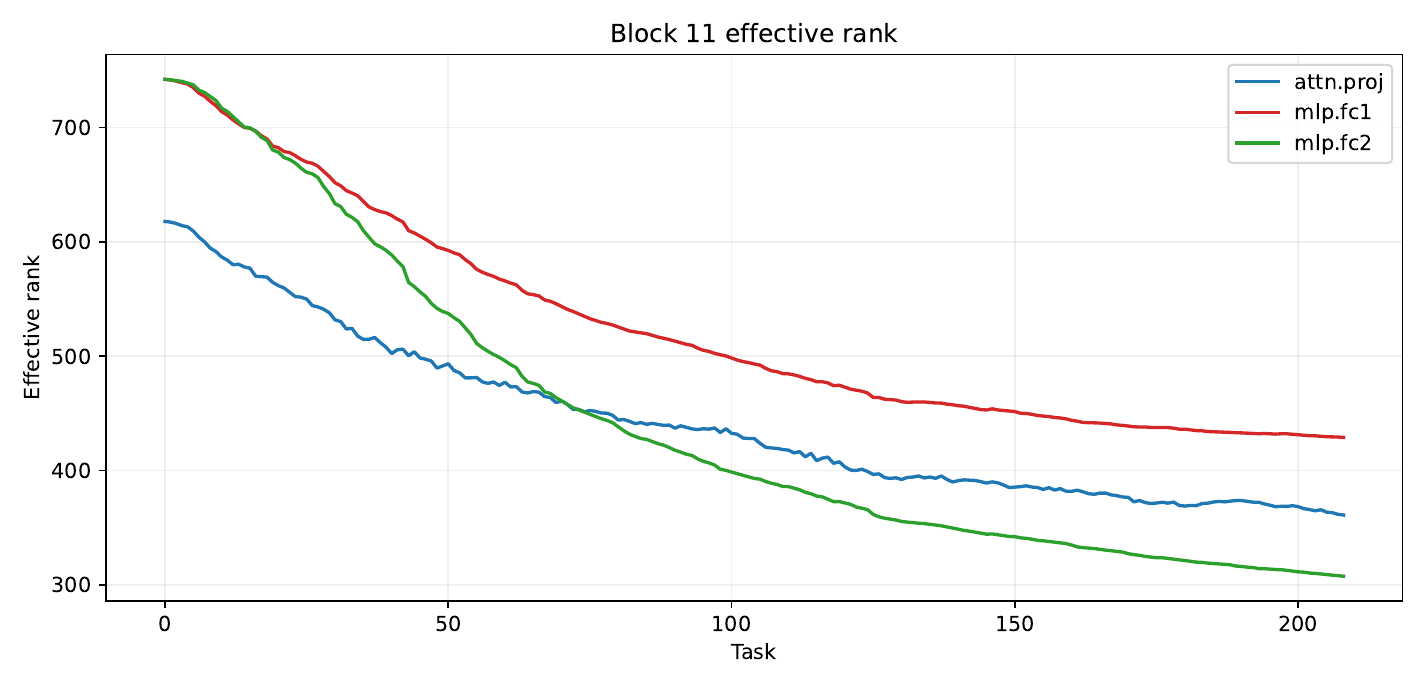}
  \includegraphics[width=0.45\textwidth]{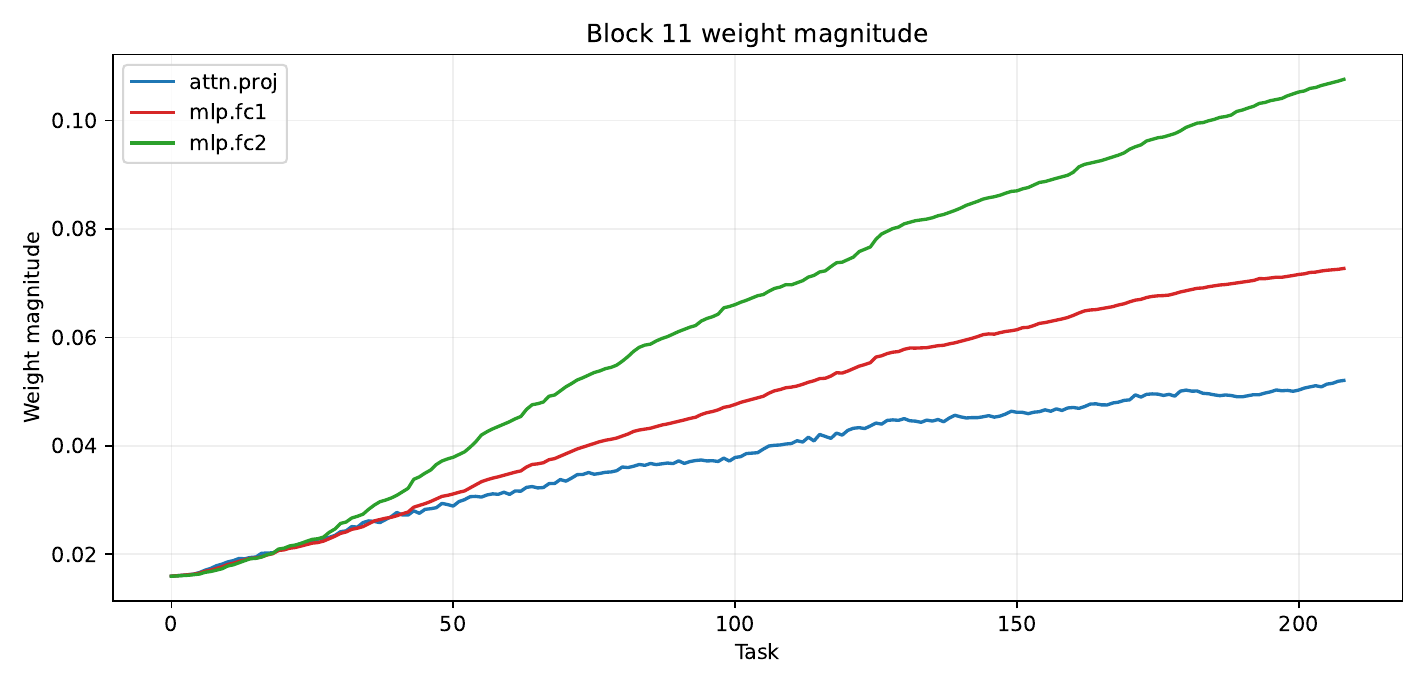}
  \caption{Effective rank (left) and weight magnitude (right) of attention projections and FFN of ViT, block 12. The blue curves represent the metrics of attention projection. The red and green curves represent the first and second FC layers in FFN.}
\end{figure}
\newpage
\subsection{Weight Metrics Across Attention Heads}
\label{app:weight2}
\begin{figure}[htbp]
  \centering
  \includegraphics[width=0.8\textwidth]{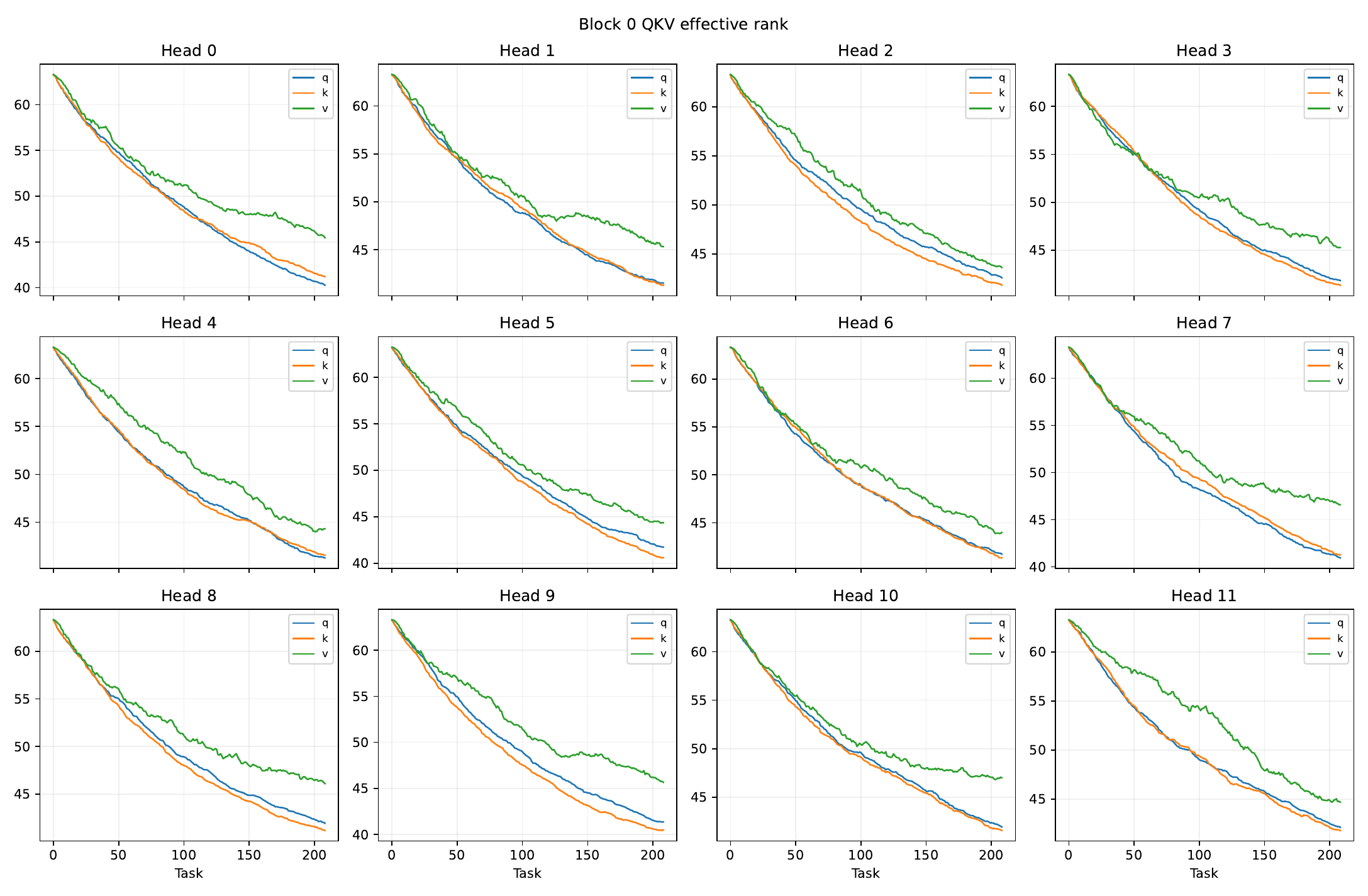}
  \caption{The effective rank of 12 attention heads in ViT, block 1. The green curve represents the $V$ matrix, which is the most unstable one. The blue and red curves represent the $Q$ and $K$, respectively.}

\end{figure}
\begin{figure}[htbp]
  \centering
  \includegraphics[width=0.8\textwidth]{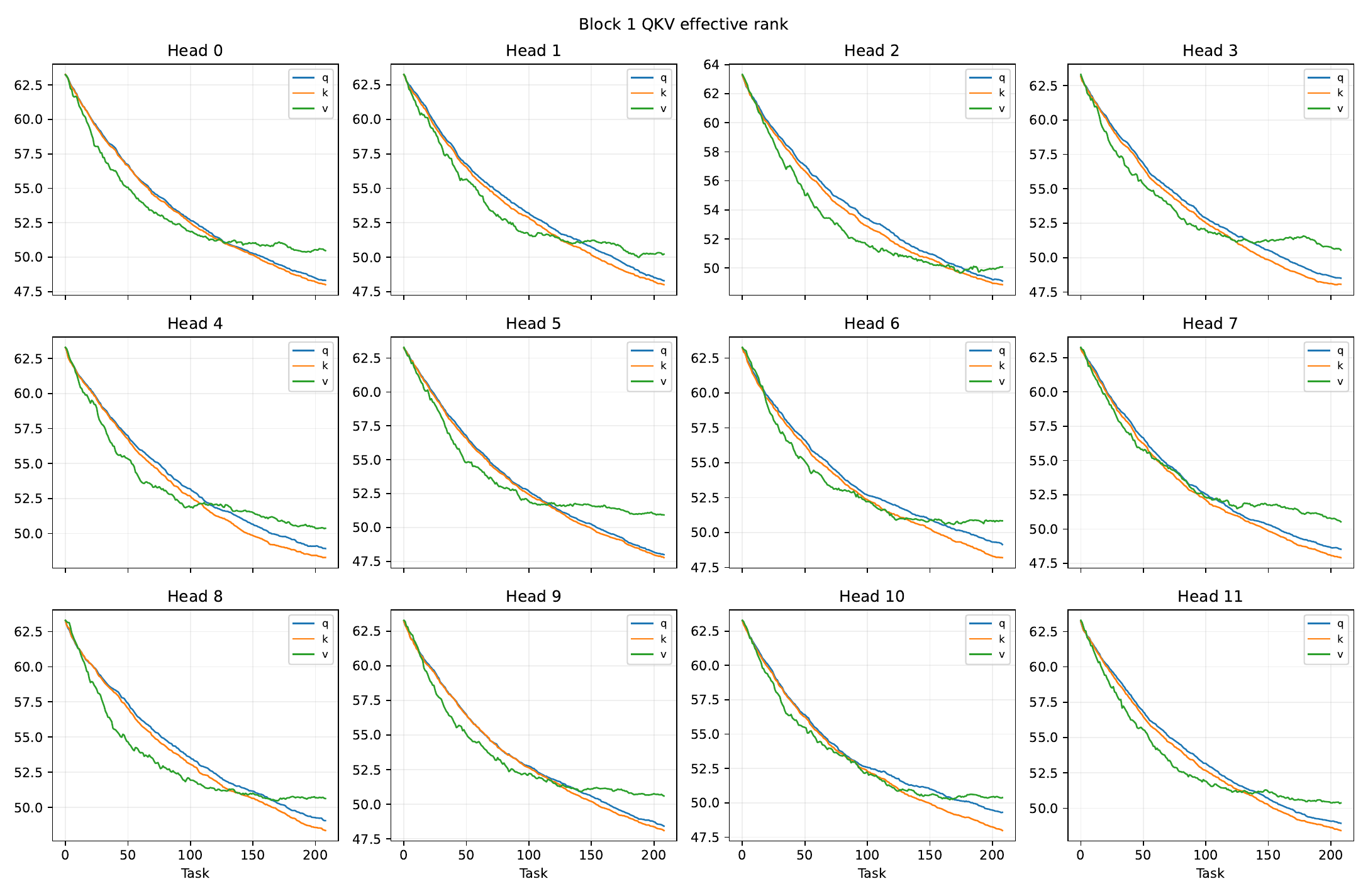}
  \caption{The effective rank of 12 attention heads in ViT, block 2. The green curve represents the $V$ matrix, which is the most unstable one. The blue and red curves represent the $Q$ and $K$, respectively.}

\end{figure}
\begin{figure}[htbp]
  \centering
  \includegraphics[width=0.8\textwidth]{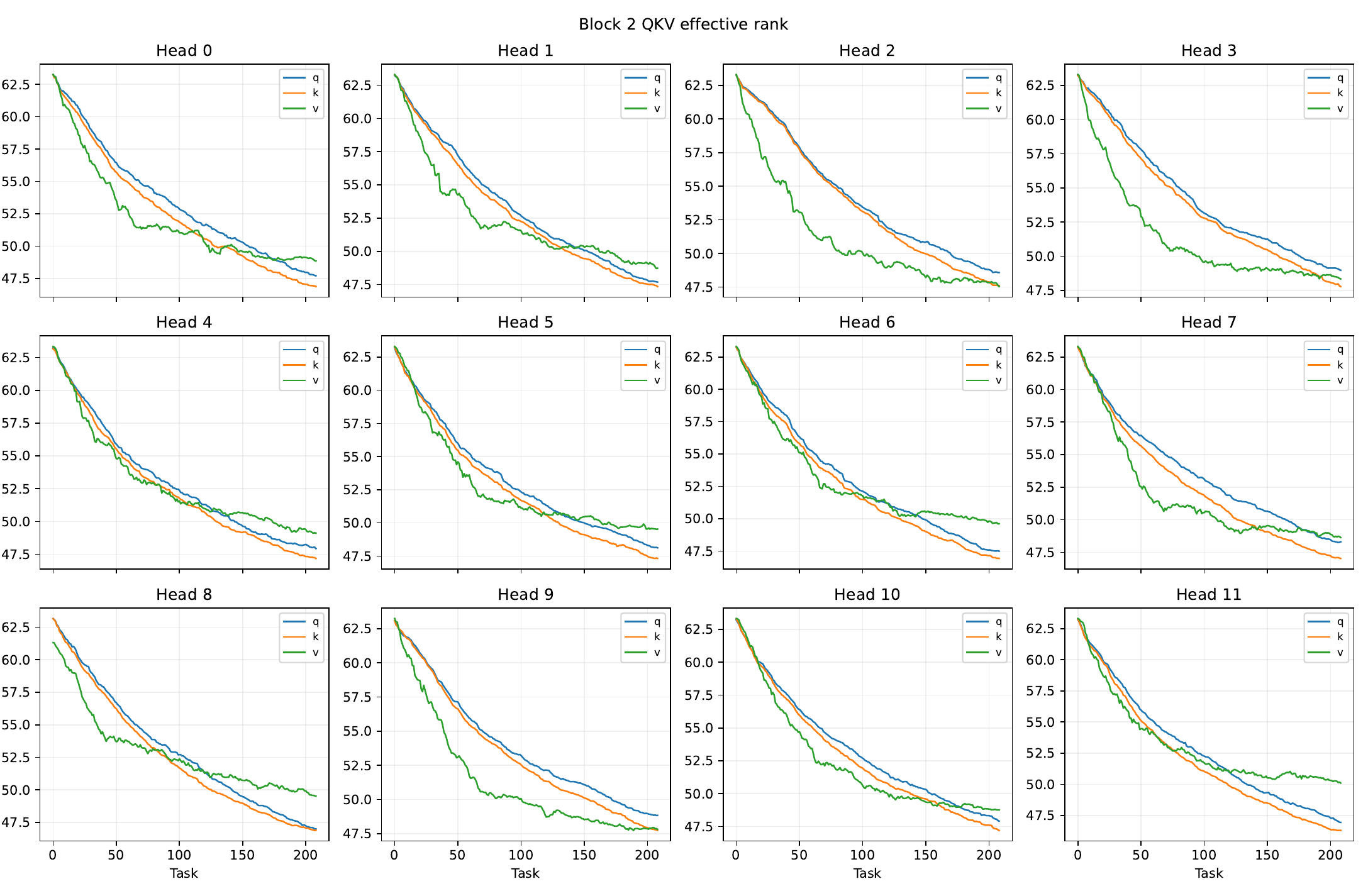}
  \caption{The effective rank of 12 attention heads in ViT, block 3. The green curve represents the $V$ matrix, which is the most unstable one. The blue and red curves represent the $Q$ and $K$, respectively.}

\end{figure}
\begin{figure}[htbp]
  \centering
  \includegraphics[width=0.8\textwidth]{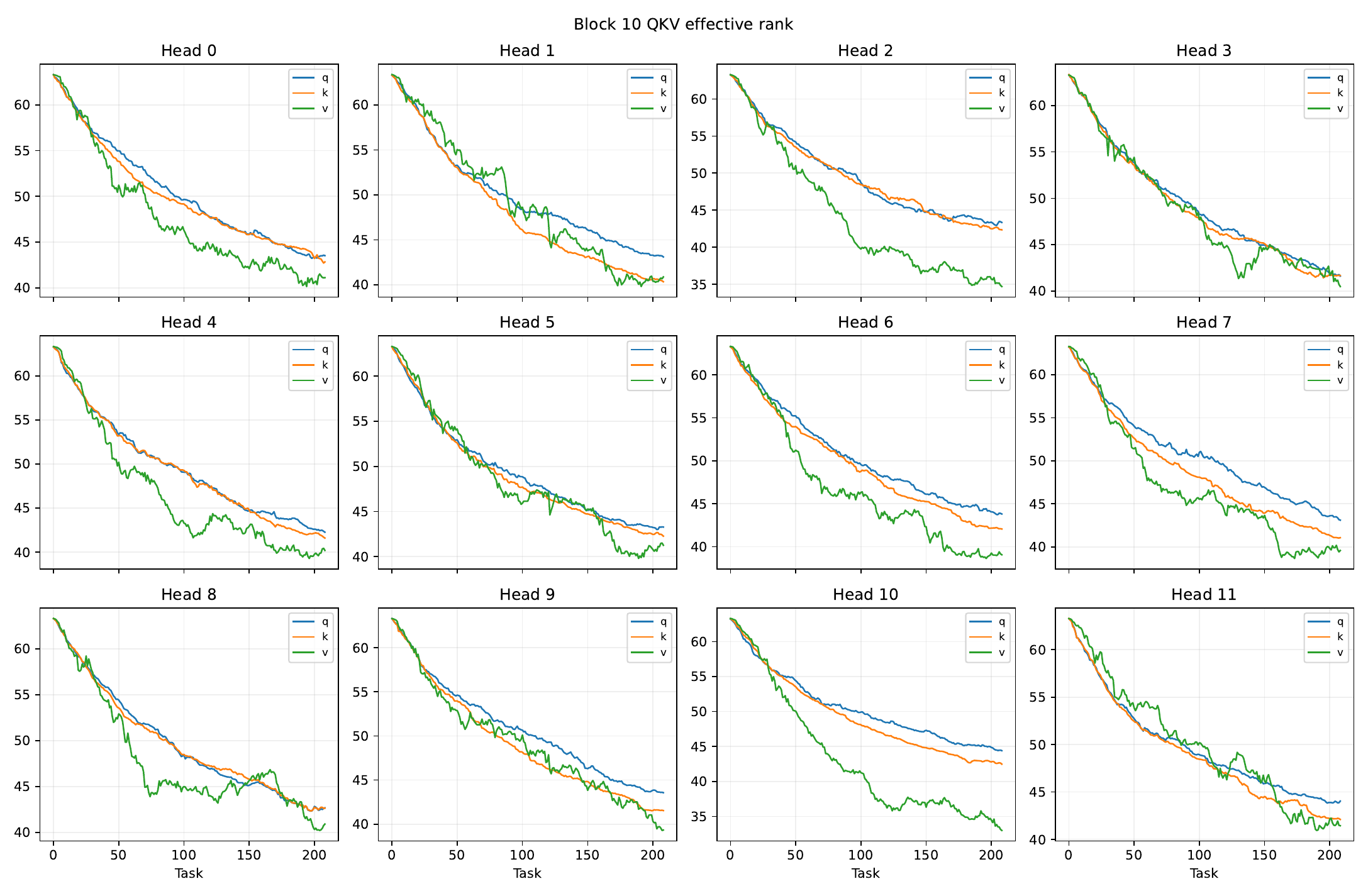}
  \caption{The effective rank of 12 attention heads in ViT, block 11. The green curve represents the $V$ matrix, which is the most unstable one. The blue and red curves represent the $Q$ and $K$, respectively.}

\end{figure}
\begin{figure}[htbp]
  \centering
  \includegraphics[width=0.8\textwidth]{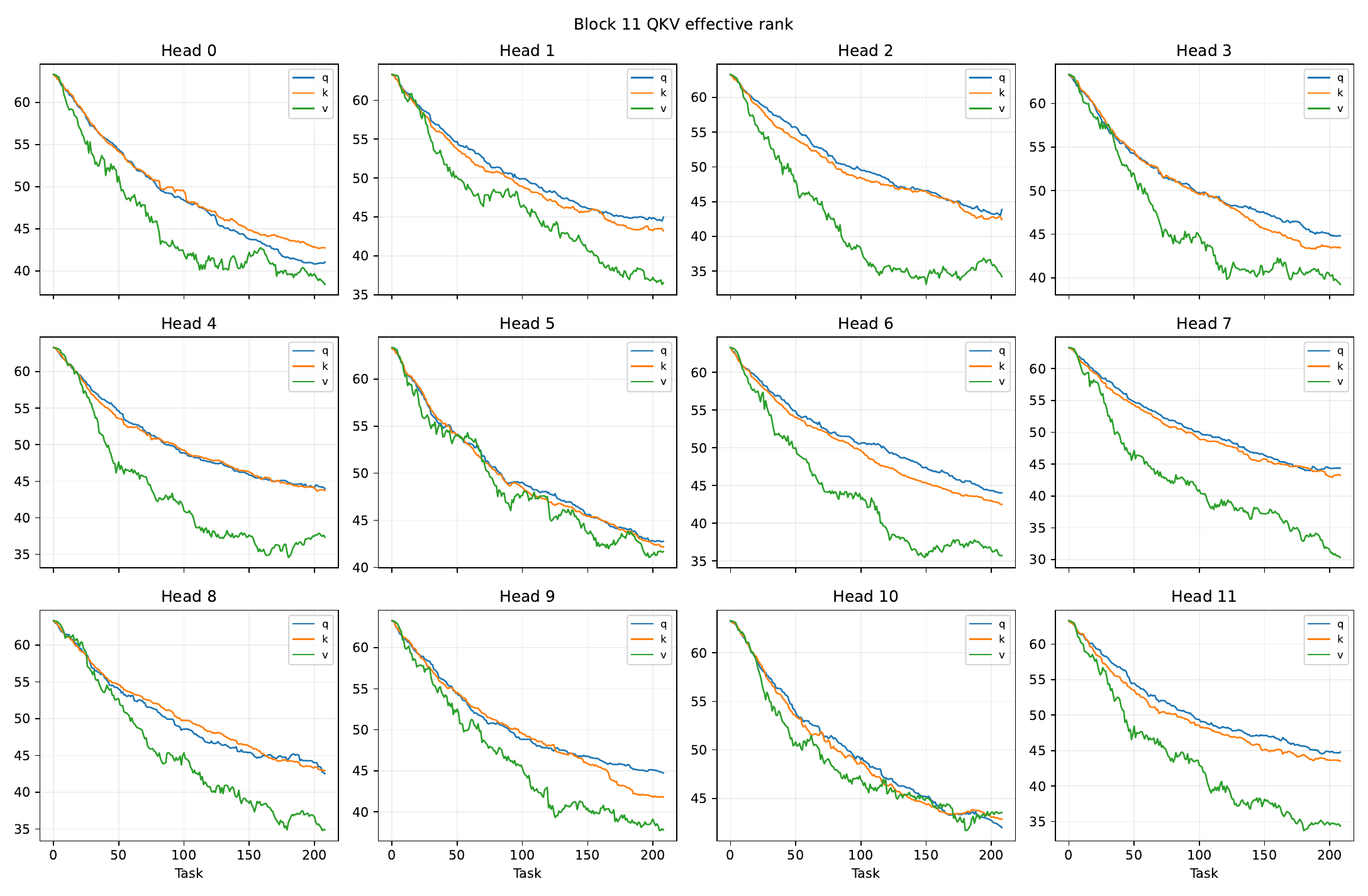}
  \caption{The effective rank of 12 attention heads in ViT, block 12. The green curve represents the $V$ matrix, which is the most unstable one. The blue and red curves represent the $Q$ and $K$, respectively.}

\end{figure}
\begin{figure}[htbp]
  \centering
  \includegraphics[width=0.8\textwidth]{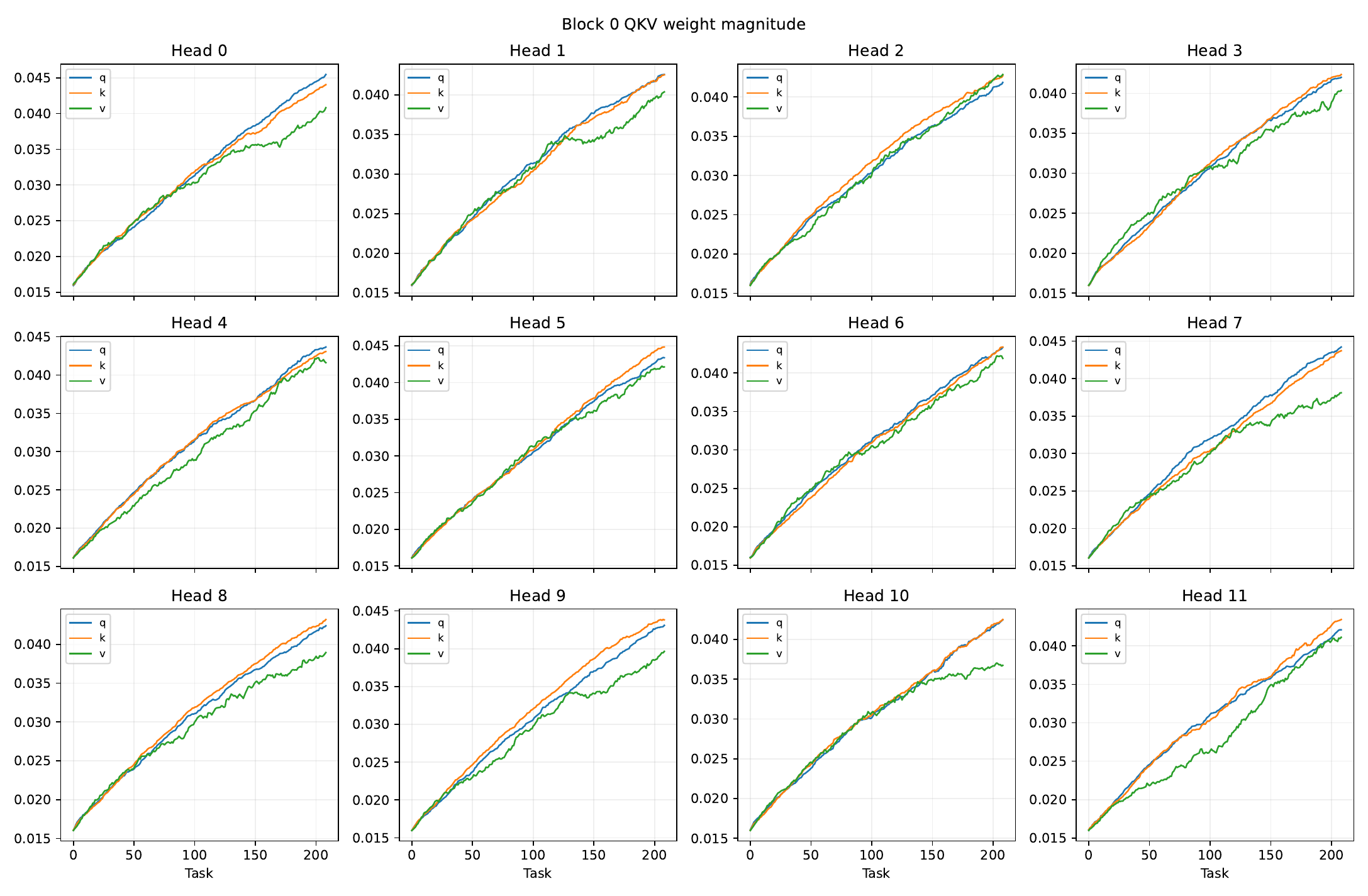}
  \caption{The weight magnitude of 12 attention heads in ViT, block 1. The green curve represents the weight magnitude of $V$ matrix. The blue and red curves represent the weight magnitude of $Q$ and $K$, respectively.}

\end{figure}
\begin{figure}[htbp]
  \centering
  \includegraphics[width=0.8\textwidth]{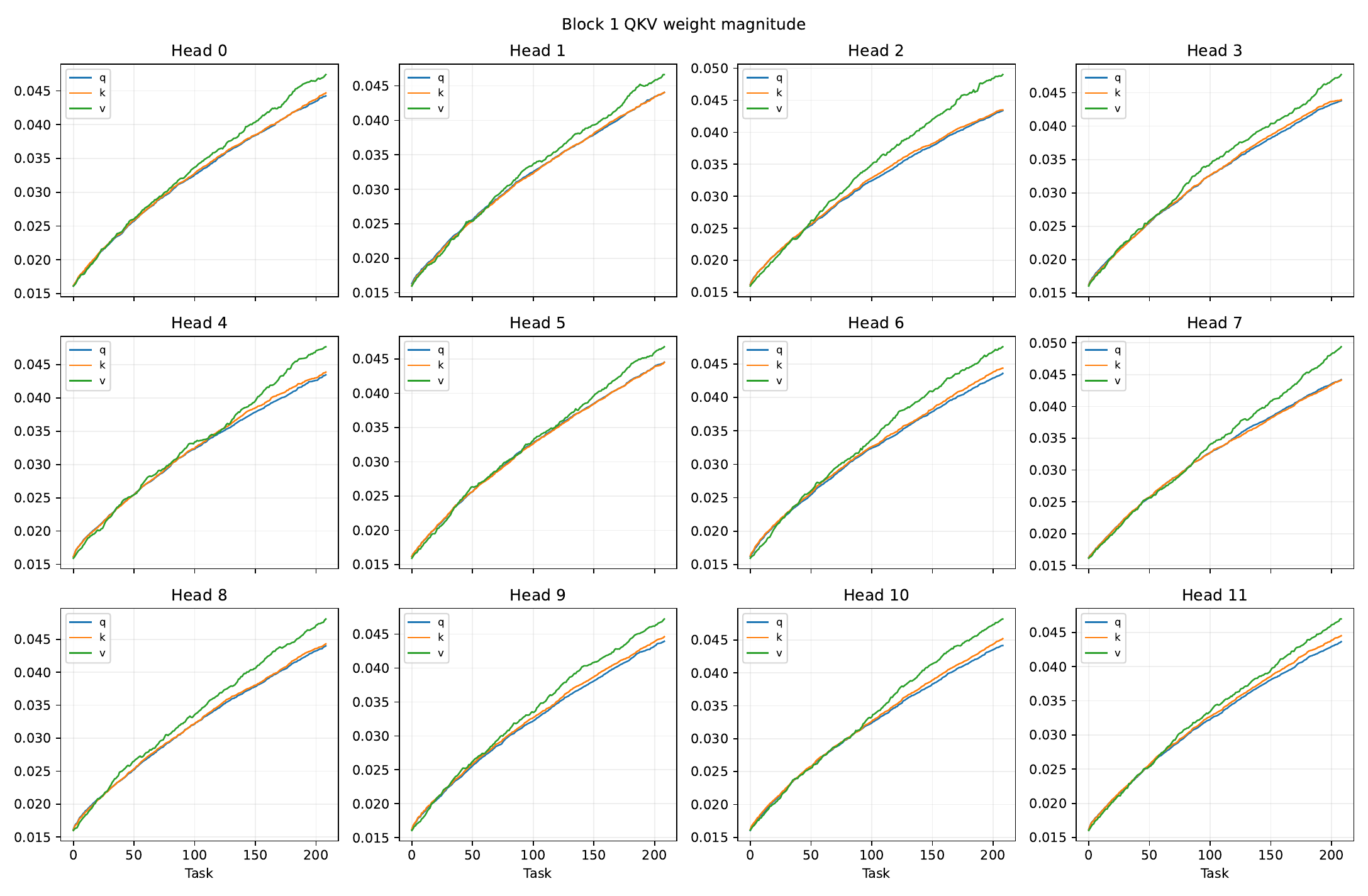}
  \caption{The weight magnitude of 12 attention heads in ViT, block 2. The green curve represents the weight magnitude of $V$ matrix. The blue and red curves represent the weight magnitude of $Q$ and $K$, respectively.}

\end{figure}
\begin{figure}[htbp]
  \centering
  \includegraphics[width=0.8\textwidth]{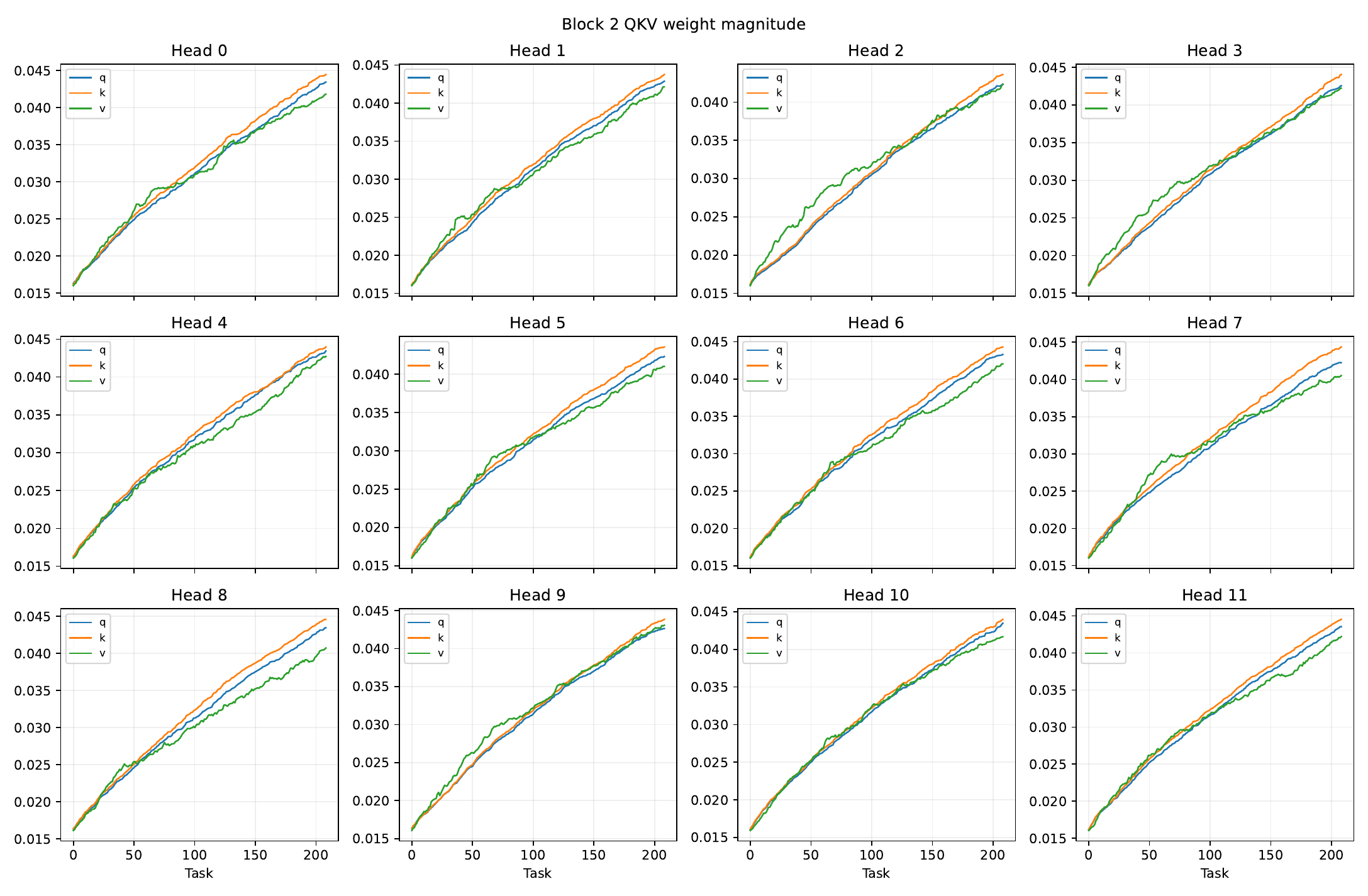}
  \caption{The weight magnitude of 12 attention heads in ViT, block 3. The green curve represents the weight magnitude of $V$ matrix. The blue and red curves represent the weight magnitude of $Q$ and $K$, respectively.}

\end{figure}
\begin{figure}[htbp]
  \centering
  \includegraphics[width=0.8\textwidth]{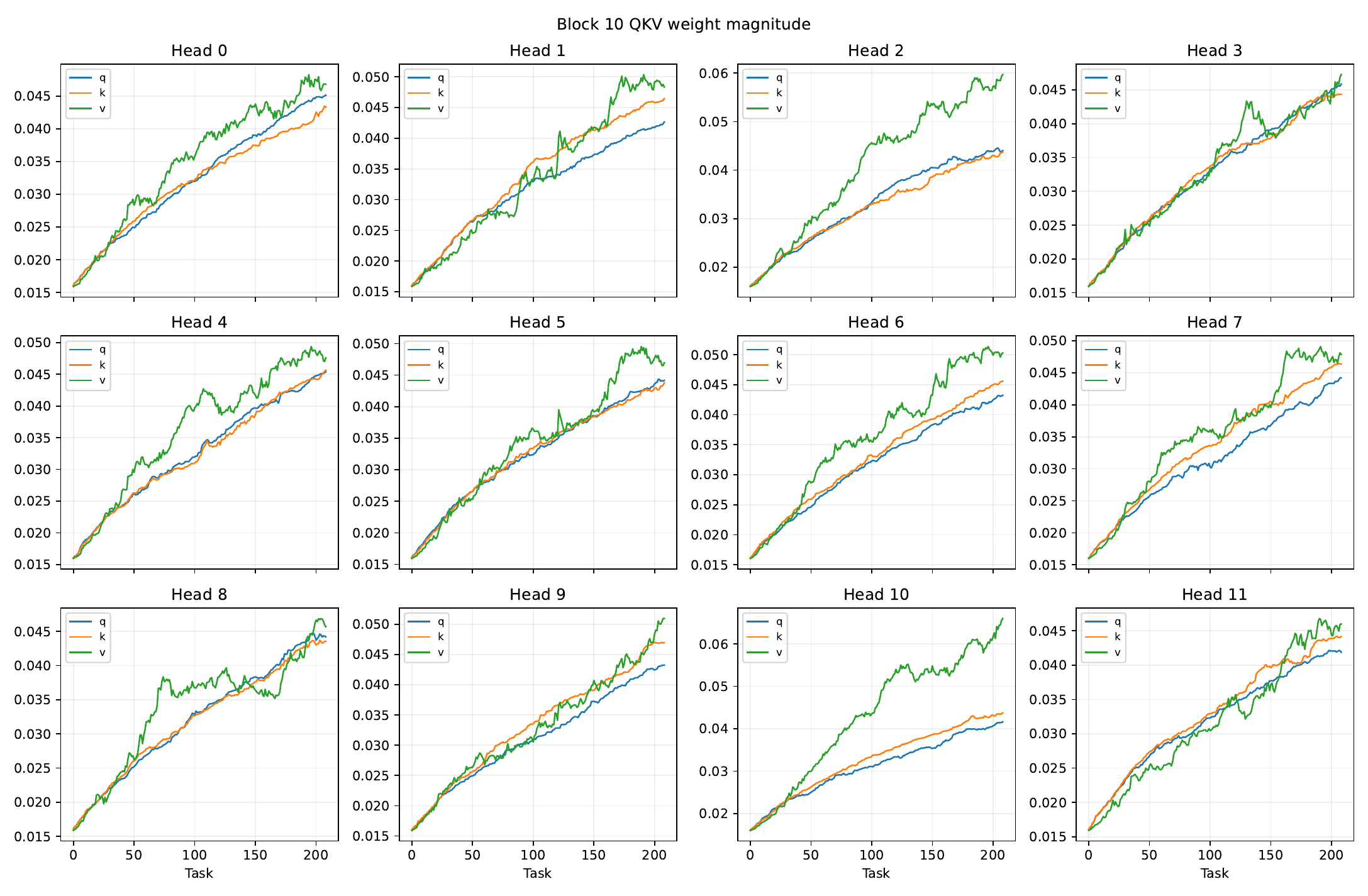}
  \caption{The weight magnitude of 12 attention heads in ViT, block 11. The green curve represents the weight magnitude of $V$ matrix. The blue and red curves represent the weight magnitude of $Q$ and $K$, respectively.}

\end{figure}
\begin{figure}[htbp]
  \centering
  \includegraphics[width=0.8\textwidth]{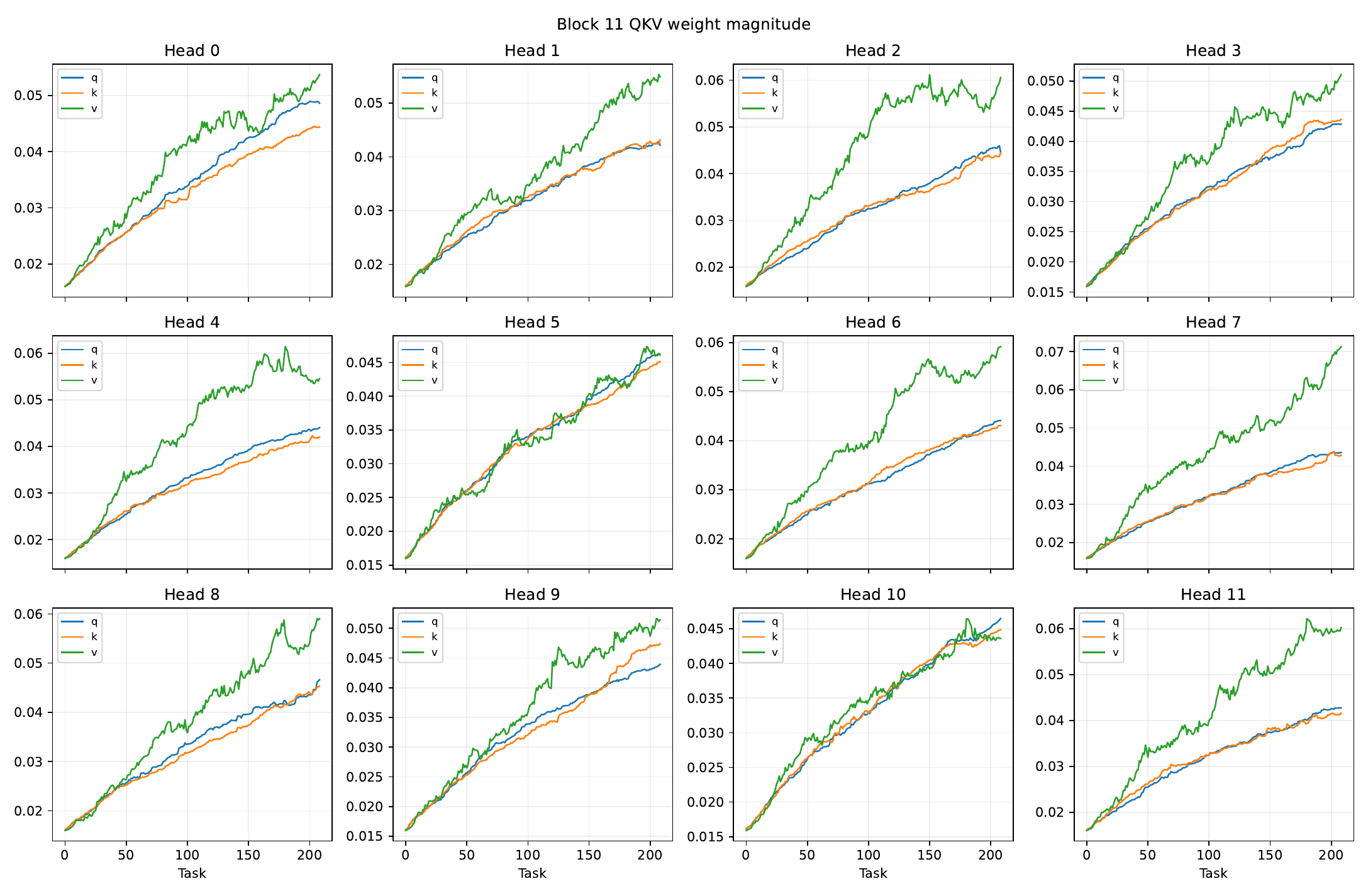}
  \caption{The weight magnitude of 12 attention heads in ViT, block 12. The green curve represents the weight magnitude of $V$ matrix. The blue and red curves represent the weight magnitude of $Q$ and $K$, respectively.}
\end{figure}

\newpage
\subsection{The Track of CLS token in Sampled Tasks}
\label{app:cls}
\begin{figure}[htbp]
  \centering
  \includegraphics[width=0.6\textwidth]{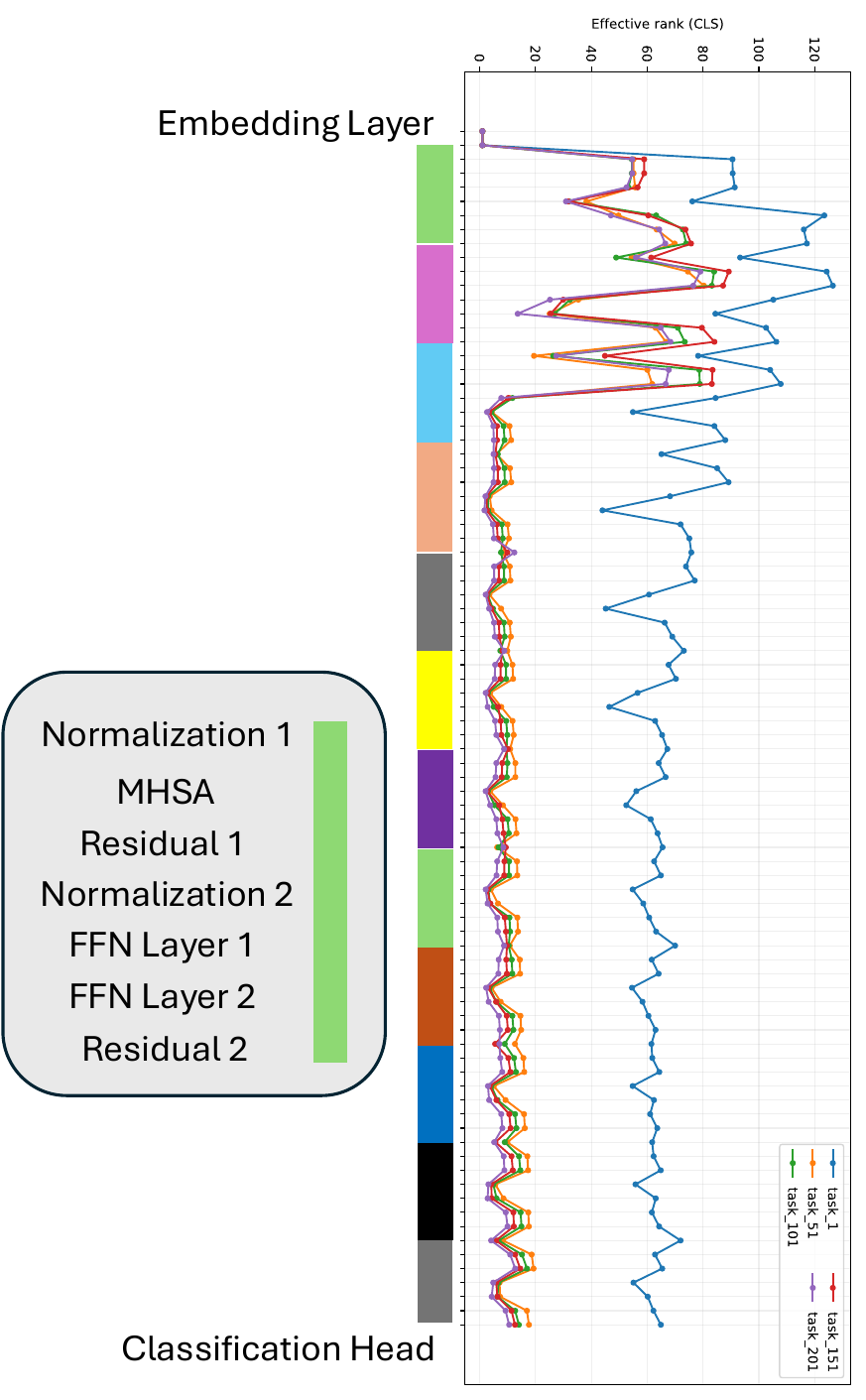}
  \caption{The effective rank of each component of cls token in ViT during continual learning. The order of components starts from embedding layers, 12 ViT blocks, and classification head. The sampled tasks are task 1 (blue), task 51 (orange), task 101 (green), task 151 (red), and task 201 (purple).}
\end{figure}

\newpage
\subsection{The Fraction of Active/Dead Units in ViT}
\label{app:fau}
\begin{figure}[htbp]
  \centering
  \includegraphics[width=0.5\textwidth]{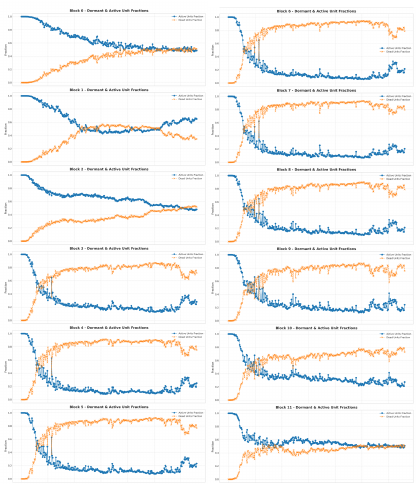}
  \caption{The FAU (blue) and FDU (orange) in activation function of FFN in ViT's 12 blocks. The significant decrease of FAU and the rise of FDU demonstrate the loss of adaptability in FFN module.}
\end{figure}

\section{Appendix}
\subsection{Performance of Plasticity Loss mitigating Method: CReLU}
\label{app:crelu}
\begin{figure}[htbp]
  \centering
  \includegraphics[width=0.7\textwidth]{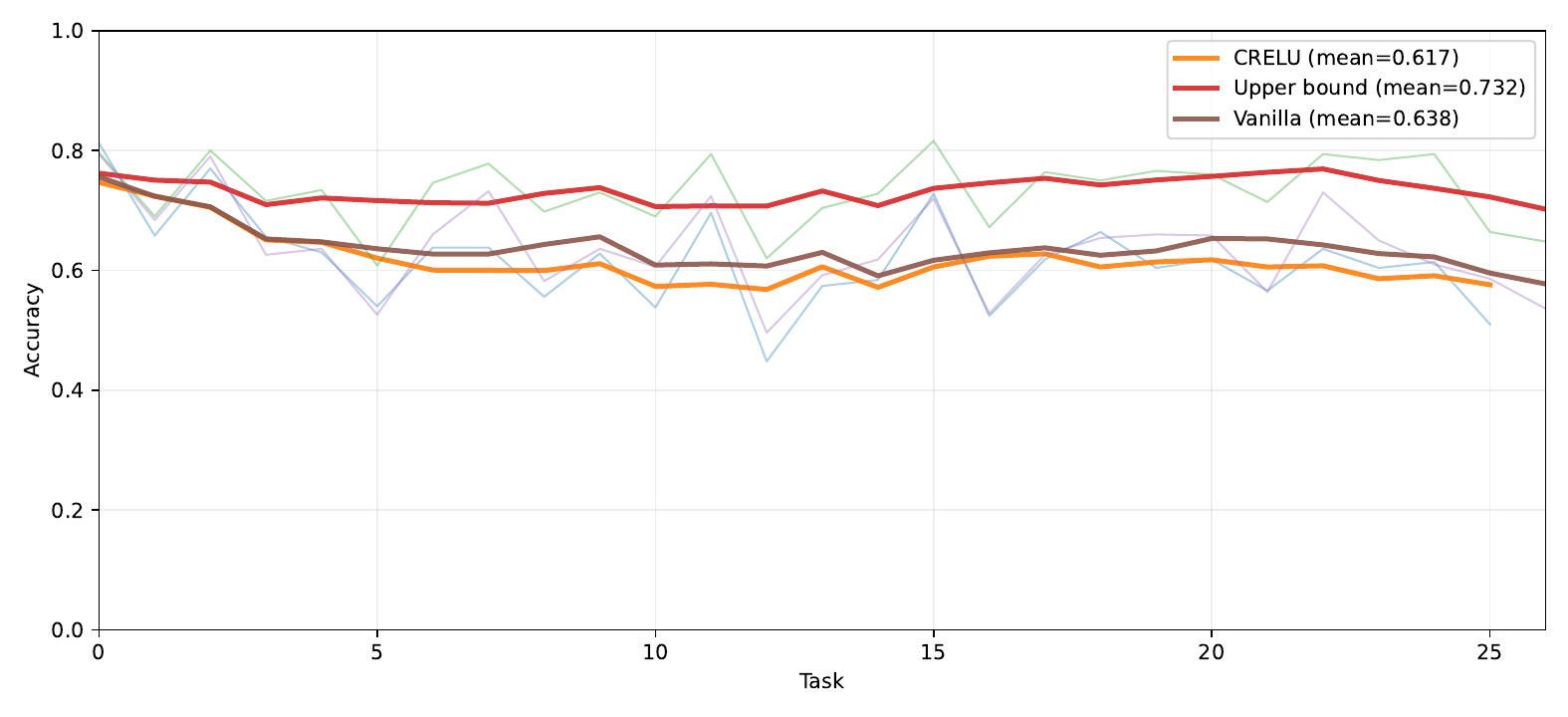}
  \caption{The accuracy curves of Upper bound (red), vanilla ViT (brown), and ViT with CReLU (orange) on first 25 tasks of stream. The AAT of CReLU is 0.617, which is less than vanilla ViT (0.638) and Upper bound (0.732), indicating the worse performance of CReLU. Moreover, the cost of CReLU is double than other approaches.}
\end{figure}

\newpage

\section{Appendix}
\subsection{The Hyperparameter Adjustment Results}
The details of the entire hyperparameters tuning of ARROW including fixed values and decay schedule. Here, $a$ annotates with $\alpha$ of ARROW. $a1e-1$ represents the value of $\alpha$ is $1\times \mathrm{e}^{-1}$. Similarly, $w20$ represents the window size of the gradient buffer is 20; $b0.9$ means the value of $\beta$ is 0.9. The followed 'rms' means the employment of RMS-like warm-up mode, whereas the SGD is used without 'rms'. 'Trac' represent the adaptive learning rate is introduced to ARROW. If the decay schedule is introduced for $\alpha$ per task, we can see symbol of 'pertask'.
\label{app:para}
\begin{table*}[h]
\centering
\caption{The accuracies of different combinations of hyperparameters for ARROW on task stream and corresponding AAT.}
\label{tab:block1}
\resizebox{\textwidth}{!}{%
\begin{tabular}{lcccccccc}
\toprule
 & a1w20b1 & a1e-1w20b01rms & a5e-1w20b1 & a5e-2w20b1 & a1e-2w20b1 & a3e-3w20b1 & a1e-3w20b1 & a1e-4w20b1 \\
\midrule
task1  & 0.595 & 0.576 & 0.632 & 0.607 & 0.603 & 0.607 & 0.595 & 0.595 \\
task2  & 0.658 & 0.659 & 0.652 & 0.648 & 0.669 & 0.648 & 0.658 & 0.658 \\
task3  & 0.639 & 0.655 & 0.600 & 0.642 & 0.651 & 0.642 & 0.639 & 0.639 \\
task4  & 0.657 & 0.667 & 0.620 & 0.661 & 0.665 & 0.661 & 0.657 & 0.657 \\
task5  & 0.609 & 0.625 & 0.602 & 0.613 & 0.602 & 0.613 & 0.609 & 0.609 \\
task6  & 0.583 & 0.606 & 0.562 & 0.627 & 0.629 & 0.627 & 0.583 & 0.583 \\
task7  & 0.510 & 0.534 & 0.488 & 0.503 & 0.537 & 0.503 & 0.510 & 0.510 \\
task8  & 0.512 & 0.528 & 0.495 & 0.524 & 0.486 & 0.524 & 0.512 & 0.523 \\
task9  & 0.630 & 0.635 & 0.621 & 0.662 & 0.638 & 0.662 & 0.630 & 0.653 \\
task10 & 0.558 & 0.556 & 0.536 & 0.558 & 0.557 & 0.558 & 0.558 & 0.590 \\
\midrule
AAT    & 0.5951 & 0.6041 & 0.5808 & 0.6045 & 0.6037 & 0.6045 & 0.5951 & 0.6017 \\
\bottomrule
\end{tabular}%
}
\end{table*}
\begin{table*}[h]
\centering
\caption{The accuracies of different combinations of hyperparameters for ARROW on task stream and corresponding AAT.}
\label{tab:block2}
\resizebox{0.9\textwidth}{!}{%
\begin{tabular}{lccccc}
\toprule
 & a1e-2w20b1rms2trac & a1e-3w20b12trac & a3e-3w20b1rms\_2trac & a5e-2w20b1rms\_2trac & a1e-3w20b0.9rms2trac \\
\midrule
task1  & 0.587 & 0.612 & 0.593 & 0.619 & 0.611 \\
task2  & 0.644 & 0.647 & 0.643 & 0.638 & 0.665 \\
task3  & 0.673 & 0.666 & 0.668 & 0.670 & 0.669 \\
task4  & 0.686 & 0.653 & 0.665 & 0.658 & 0.672 \\
task5  & 0.644 & 0.651 & 0.643 & 0.646 & 0.654 \\
task6  & 0.644 & 0.639 & 0.644 & 0.649 & 0.649 \\
task7  & 0.578 & 0.582 & 0.590 & 0.527 & 0.587 \\
task8  & 0.586 & 0.588 & 0.583 & 0.507 & 0.588 \\
task9  & 0.662 & 0.692 & 0.689 & 0.658 & 0.688 \\
task10 & 0.613 & 0.631 & 0.600 & 0.574 & 0.613 \\
\midrule
AAT    & 0.6317 & 0.6361 & 0.6318 & 0.6146 & 0.6396 \\
\bottomrule
\end{tabular}%
}
\end{table*}

\begin{table}[h]
\centering
\caption{The accuracies of different combinations of hyperparameters for ARROW on task stream and corresponding AAT.}
\label{tab:block3}
\resizebox{0.7\textwidth}{!}{%
\begin{tabular}{lcccc}
\toprule
 & c1e-2-3w20b1rmspertask & c1e-2-3w20b1rms2trcapertask & trac & baseline \\
\midrule
task1  & 0.620 & 0.614 & 0.603 & 0.632 \\
task2  & 0.669 & 0.658 & 0.653 & 0.652 \\
task3  & 0.655 & 0.677 & 0.656 & 0.606 \\
task4  & 0.647 & 0.680 & 0.683 & 0.617 \\
task5  & 0.621 & 0.658 & 0.656 & 0.575 \\
task6  & 0.614 & 0.646 & 0.648 & 0.542 \\
task7  & 0.541 & 0.570 & 0.572 & 0.463 \\
task8  & 0.503 & 0.598 & 0.586 & 0.493 \\
task9  & 0.646 & 0.679 & 0.691 & 0.630 \\
task10 & 0.562 & 0.605 & 0.610 & 0.511 \\
\midrule
AAT    & 0.6078 & 0.6385 & 0.6358 & 0.5721 \\
\bottomrule
\end{tabular}%
}
\end{table}
\begin{table*}[h]
\centering
\caption{The accuracies of different combinations of hyperparameters for ARROW on task stream and corresponding AAT.}
\label{tab:block41}
\resizebox{0.7\textwidth}{!}{%
\begin{tabular}{lccccc}
\toprule
 & a1w20b0.8 & a5e-1w20b0.9 & a1e-2w20b0.5 & a3e-3w20b1rms & a1e-3w20b0.9 \\
\midrule
task1  & 0.595 & 0.632 & 0.607 & 0.576 & 0.595 \\
task2  & 0.658 & 0.652 & 0.648 & 0.659 & 0.658 \\
task3  & 0.639 & 0.600 & 0.642 & 0.655 & 0.639 \\
task4  & 0.657 & 0.620 & 0.661 & 0.667 & 0.657 \\
task5  & 0.609 & 0.607 & 0.613 & 0.625 & 0.609 \\
task6  & 0.583 & 0.563 & 0.627 & 0.606 & 0.583 \\
task7  & 0.510 & 0.480 & 0.503 & 0.534 & 0.510 \\
task8  & 0.512 & 0.497 & 0.524 & 0.528 & 0.512 \\
task9  & 0.630 & 0.610 & 0.662 & 0.635 & 0.630 \\
task10 & 0.558 & 0.541 & 0.558 & 0.556 & 0.558 \\
\midrule
AAT    & 0.5951 & 0.5802 & 0.6045 & 0.6041 & 0.5951 \\
\bottomrule
\end{tabular}%
}
\end{table*}

\begin{table*}[h]
\centering
\caption{The accuracies of different combinations of hyperparameters for ARROW on task stream and corresponding AAT.}
\label{tab:block42}
\resizebox{0.6\textwidth}{!}{%
\begin{tabular}{lcccc}
\toprule
 & a5e-1w30b1 & a1e-2w20b1rms & a1e-3w20b0.9rms & a1e-3w10b1 \\
\midrule
task1  & 0.595 & 0.576 & 0.576 & 0.634 \\
task2  & 0.658 & 0.659 & 0.659 & 0.663 \\
task3  & 0.639 & 0.655 & 0.655 & 0.635 \\
task4  & 0.657 & 0.677 & 0.667 & 0.654 \\
task5  & 0.609 & 0.625 & 0.625 & 0.587 \\
task6  & 0.583 & 0.606 & 0.606 & 0.575 \\
task7  & 0.510 & 0.534 & 0.534 & 0.489 \\
task8  & 0.512 & 0.528 & 0.528 & 0.489 \\
task9  & 0.630 & 0.635 & 0.635 & 0.630 \\
task10 & 0.558 & 0.556 & 0.556 & 0.536 \\
\midrule
AAT    & 0.5951 & 0.6051 & 0.6041 & 0.5850 \\
\bottomrule
\end{tabular}%
}
\end{table*}
\begin{table}[h]
\centering
\caption{The accuracies of different combinations of hyperparameters for ARROW on task stream and corresponding AAT.}
\label{tab:block43}
\resizebox{0.35\textwidth}{!}{%
\begin{tabular}{lcc}
\toprule
 & a1e-3w30b1 & a1e-2w20b0.3rms \\
\midrule
task1  & 0.632 & 0.576 \\
task2  & 0.648 & 0.659 \\
task3  & 0.633 & 0.655 \\
task4  & 0.610 & 0.667 \\
task5  & 0.595 & 0.625 \\
task6  & 0.579 & 0.606 \\
task7  & 0.505 & 0.534 \\
task8  & 0.510 & 0.528 \\
task9  & 0.634 & 0.635 \\
task10 & 0.529 & 0.556 \\
\midrule
AAT    & 0.5875 & 0.6051 \\
\bottomrule
\end{tabular}%
}
\end{table}
\label{app:results}
\end{document}